\documentclass[letterpaper, 10 pt, conference]{ieeeconf}  
\pdfoutput=1

\IEEEoverridecommandlockouts                              

\usepackage{graphics} 
\usepackage{epsfig} 
\usepackage{mathptmx} 
\usepackage{times} 
\usepackage{amsmath} 
\usepackage{amssymb}  
\usepackage{subcaption}
\usepackage{gensymb}
\usepackage{xcolor}
\usepackage{booktabs}
\usepackage{cite}
\usepackage{algorithm}
\usepackage{algorithmic}
\usepackage{comment}
\usepackage{booktabs}
\usepackage{graphicx}

\title{\LARGE \bf
SceNDD: A Scenario-based Naturalistic Driving Dataset}

\author{Avinash Prabu, Nitya Ranjan, Lingxi Li*, Renran Tian, Stanley Chien, Yaobin Chen, and Rini Sherony
\thanks{This work was supported in part by the Toyota Collaborative Safety Research Center (CSRC).}
\thanks{A. Prabu, N. Ranjan, L. Li, S. Chien, and Y. Chen are with the Transportation and Autonomous Systems Institute (TASI), also with the Department of Electrical and Computer Engineering, Indiana University-Purdue University Indianapolis. Emails: {\tt\small \{aprabu, niranjan, ll7, schien, ychen\}@iupui.edu.}}%
\thanks{R. Tian is with the Transportation and Autonomous Systems Institute (TASI), also with the Department of Computer Information and Graphics Technology, Indiana University-Purdue University Indianapolis. Email:         {\tt\small rtian@iupui.edu.}}%
\thanks{R. Sherony is with the Collaborative Safety Research Center (CSRC), Toyota Motor
North America, Ann Arbor, MI, USA. Email:         {\tt\small rini.sherony@toyota.com.}}%
\thanks{*Corresponding author.}%
}
\begin{document}

\maketitle
\thispagestyle{empty}
\pagestyle{empty}

\begin{abstract}

In this paper, we propose SceNDD: a scenario-based naturalistic driving dataset that is built upon data collected from an instrumented vehicle in downtown Indianapolis. The data collection was completed in 68 driving sessions with different drivers, where each session lasted about 20--40 minutes. The main goal of creating this dataset is to provide the research community with real driving scenarios that have diverse trajectories and driving behaviors. The dataset contains ego-vehicle's waypoints, velocity, yaw angle, as well as non-ego actor's waypoints, velocity, yaw angle, entry-time, and exit-time. Certain flexibility is provided to users so that actors, sensors, lanes, roads, and obstacles can be added to the existing scenarios. We used a Joint Probabilistic Data Association (JPDA) tracker to detect non-ego vehicles on the road. We present some preliminary results of the proposed dataset and a few applications associated with it. The complete dataset is expected to be released by early 2023. 

\end{abstract}

\section{INTRODUCTION}

The development of connected and automated vehicles (CAVs) technology has been growing at a fast pace in the last decade \cite{Libook}, \cite{TIV}. Motion planning is one of the most important aspects for CAVs. Thus, there is an urgent need for developing efficient algorithms as well as their verification and validation methods using benchmark datasets. High-quality datasets enable researchers to  virtually access, visualize, simulate, and analyze real-life urban driving scenarios, which greatly save their time and energy. In this paper, we develop SceNDD: a scenario-based naturalistic driving dataset that is built upon data collected from an instrumented vehicle in downtown Indianapolis. The data collection was completed in 68 driving sessions with different drivers, where each session lasts 20--40 minutes. The main goal of creating this dataset is to provide the research community with real driving scenarios that have diverse driving trajectories and behaviors to develop efficient motion planning and path following algorithms.

A multi-model dataset typically needs a combination of sensors for capturing data. Cameras are used for recording images/videos and the features of interest can be extracted and analyzed using image processing or computer vision algorithms. LiDAR data are used for 3D localization. Inertial measurement unit (IMU)  and GPS devices are also popular to get accurate position, trajectory, velocity, and acceleration of the ego-vehicle.  In literature, datasets such as Waymo Open Dataset\cite{Waymo}, nuScenes\cite{nuscenes}, LYFT\cite{LYFT}, KITTI\cite{Fritsch2013ITSC}, Argoverse\cite{Argoverse}, and ApolloScape \cite{Apolloscape} have been created using a combination of sensors mentioned above. They are popular benchmark datasets for the research community to use and develop efficient and effective algorithms for CAVs. 

The nuScenes\cite{nuscenes} dataset is one of the most commonly used benchmark datasets for CAVs. It includes around 6 hours of driving data captured by front-facing stereo cameras, LiDAR, and GPS/IMU sensors. The LYFT \cite{LYFT} and KITTI \cite{Fritsch2013ITSC} datasets are similar, with difference in the perception environments and data sizes. These datasets are excellent for developing perception algorithms. However, they do not focus on capturing motion features such as driving scenarios and driving behavior. 

Recently, multiple motion datasets have been proposed, which include various driving scenes, tracked objects, moving trajectories, and time segments. A notable dataset is the Waymo Open dataset \cite{Waymo}, which includes motion prediction, interaction prediction, and occupancy/flow prediction. It is composed of 574 hours of naturalistic driving data, collected in different major cities in the U.S. Argoverse\cite{Argoverse} is also a dataset that focuses on certain driving behaviors, such as turning at an intersection and changing lanes. The Interaction dataset\cite{interactiondataset} is collected via drone footage and contains naturalistic motion of various traffic agents in multiple driving scenarios in various countries, which has diverse and complex behaviors of vehicles at intersections and roundabouts. However, this dataset does not have 3D state estimates that our dataset includes. The ApolloScape dataset\cite{Apolloscape} also includes trajectories of different traffic agents but does not have trajectory of the ego vehicle. Datasets such as the Stanford Drone Dataset\cite{StanfordDroneDataset}, NGSIM\cite{NGSIM}, ETH\cite{ETH}, UCY\cite{UCY}, Town Center\cite{TownCenter} are popular in motion forecasting but do not focus on the driving environment and are much smaller in size compared to our dataset. 

\begin{table*}[t]
\centering
\caption{Comparison of various autonomous driving datasets}
\begin{tabular}{|c|c|c|c|}
\toprule
Name                    & Task                   & Focus       & Environment                   \\ \midrule
nuscenes                & Perception, Prediction & Varied      & Urban Streets                 \\
Waymo Open Dataset      & Perception, Prediction & Varied      & Varied                        \\
LYFT                    & Perception, Planning   & Varied      & Varied                        \\
KITTI                   & Perception             & Pedestrians & Sidewalk                      \\
Argoverse Forecasting   & Prediction             & Varied      & Urban Streets                 \\
AppolloScape Trajectory & Prediction             & Varied      & Urban Streets                 \\
The Interaction Dataset      & Prediction                                & Vehicles                      & Ramps, Roundabouts,  Intersections           \\
Stanford Drone Dataset       & Perception                                & Pedestrians                   & Pedestrian only streets, Pedestrian Crossing \\
NGSIM                   & Prediction             & Pedestrians & Highway                       \\
ETH                     & Prediction             & Pedestrians & Pedestrian Crossing, Sidewalk \\
UCY                     & Prediction             & Pedestrians & Pedestrian Crossing, Sidewalk \\
Town Center                  & Prediction                                & Pedestrians                   & Sidewalk, Pedestrian only Streets            \\ \midrule
\multicolumn{1}{|l|}{SceNDD} & \multicolumn{1}{l|}{Prediction, Planning, Flexibility} & \multicolumn{1}{l|}{Vehicles} & \multicolumn{1}{l|}{Varied}                  \\ \bottomrule
\end{tabular}%

\label{table:1}
\end{table*}

Our dataset includes naturalistic driving data of 68 driving sessions with different drivers, where each session lasts 20--40 minutes. One unique feature is that it provides flexibility for users to add/remove targets in the simulated environment and customize the environment to their needs. The development is based on the Matlab, which allows the addition of different sensors (e.g., radar, LiDAR, ultrasonic sensors, etc.) to the ego vehicle, so that it is adaptive to different configurations of the ego-vehicle. Another attribute of our dataset is the time horizon of prediction. Since all the data is broken down into scenarios, the data can be easily broken down into timestamps of different history and horizon time. The comparison of various autonomous driving dataset is shown in Table \ref{table:1}.

To the best of our knowledge, only a few existing datasets provide scenarios with flexibility in diverse real-time driving. In this paper, we propose creating a dataset that involves a diverse mix of driving data collected using different drivers around downtown Indianapolis. The main contributions of this paper are summarized as follows: 
\begin{itemize}
    \item Create a comprehensive dataset that has 68 driving sessions around downtown Indianapolis, with each session being 20-40 minutes.
    \item Process and fuse data from different sensors. In this paper, since only GPS and the LiDAR data are used for processing, a simple time synchronization of the sensors is sufficient. In the future versions of the dataset, a camera-LiDAR sensor fusion will be adopted. 
    \item Develop a pipeline to generate driving scenarios from recorded vehicle data.
\end{itemize}


\section{Data Acquisition System}
\begin{figure}[h!]
    \centering
    \centering
    \includegraphics [scale = 0.7] {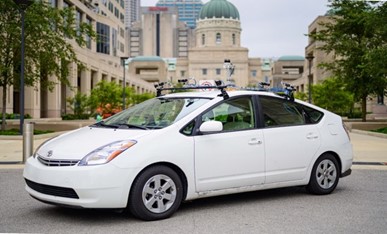}
    \caption{The instrumented vehicle for data collection.}
    \label{fig:ExperimentVehicle}
\end{figure}

In this section, we briefly describe the development of vehicle-based data acquisition system (DAS), which includes six cameras to cover all 360-degree angles, one 64-beam 360-degree Ouster LiDAR, a Reach Emlid GPS, and a desktop computer. The instrumented vehicle is shown in Fig. \ref{fig:ExperimentVehicle}.

\subsection{Cameras}

\begin{figure}[h!]
    \centering
    \centering
    \includegraphics [scale = 0.7] {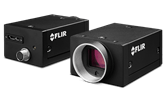}
    \caption{FLIR Grasshopper-3 camera.}
    \label{fig:camera}
\end{figure}

The main features considered while selecting a camera were resolution, frames per second (FPS), shutter type, connector type, color coding, external hardware synchronization, data transfer speed, compatibility with ROS operating system, and cost. FLIR Grasshopper-3 cameras are selected, which are capable of providing a resolution of $2048 \times 2048$, 1 to 90 FPS, and programmable exposure and shutter speed. The camera also has a global shutter, which reduces blur due to fast motion. The camera has a CMOS sensor to facilitate noise reduction and has a GPIO port capable of bi-directional synchronization with the other sensors. We used two types of lenses, 95$^{\circ}$ and 43$^{\circ}$. The wider-angle lens was used to cover the sides of the vehicle, and the narrower-angle lens was used to cover the front and back. The camera used is shown in Fig. \ref{fig:camera}.

\subsection{LiDAR}
\begin{figure}[h!]
    \centering
    \centering
    \includegraphics [scale = 0.5] {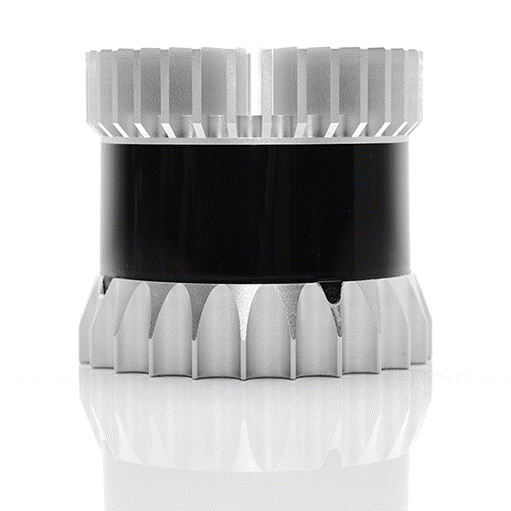}
    \caption{Ouster OS-1 64-beam LiDAR.}
    \label{fig:Lidar}
\end{figure}

The Ouster OS-1 64-beam medium-range LiDAR was used in our data collection. The OS-1 can generate 1,310,720 points at a frequency of 10 Hz, has a horizontal field of 360 degrees, and a 45 degrees vertical field of view. The LiDAR has a maximum distance range of 120m with an accuracy of +/- 5cm for Lambertian targets. OS-1 LiDAR also comes with a sensor interface that allows for transmitting data over Ethernet using UDP. The sensor interface also has a multi-purpose I/O port, which can be programmed to synchronize signals with cameras. The LiDAR is shown in Fig. \ref{fig:Lidar}.

\subsection{GPS}
\begin{figure}[h!]
    \centering
    \centering
    \includegraphics [scale = 0.2] {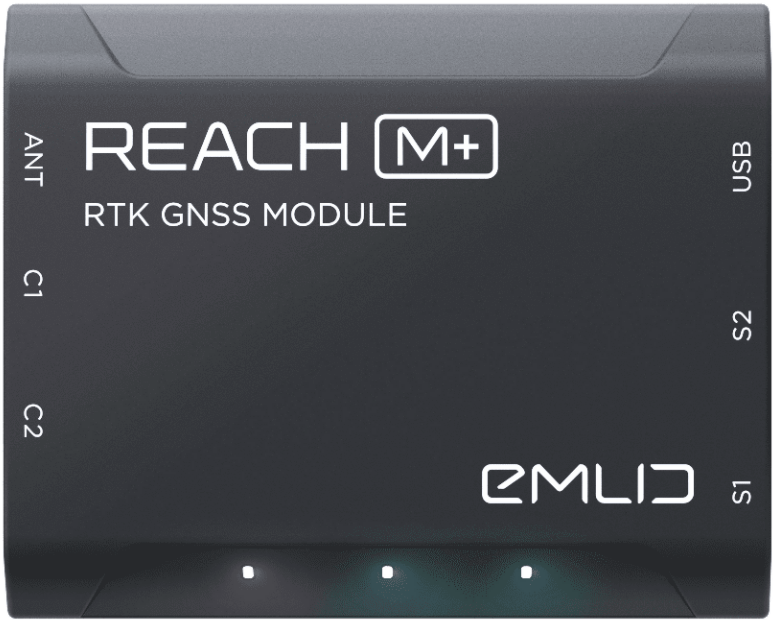}
    \caption{Reach Emlid M+ RTK GPS.}
    \label{fig:GPS}
\end{figure}

The Real-time kinematic (RTK) GPS module with base correction is utilized to detect and track the instrumented vehicle's position and motion profile. The system has three key components: 1) Emlid Reach M+ RTK GNSS Module as a rover, 2) Tallysman Global Navigation Satellite System (GNSS) Antenna, and 3) INDOT-INCORS, a network of GNSS base collection systems operated by the Indiana Department of Transportation. The base station receives a signal from GPS satellites, creates a correction factor, and then sends the correction signal via WI-FI to the rover. Our system uses a 4G hotspot as the WI-FI for the communication to the base station. The Reach M+ RTK GNSS Module has a 5GB internal storage and can provide precise navigation and mapping. The GPS module is shown in Fig. \ref{fig:GPS}. 

\subsection{PC}
The computer we use runs on Linux 4.15.0-123-generic (x86-64) kernel and is installed with Ubuntu 20.04 LTS operating system. The system hosts an 8GB solid-state drive memory. The processor used is an Intel® Xeon(R) W-2123 CPU @ 3.60GHz × 8. The system also contains AMD Radeon Pro WX2100 (POLARIS12/DRM 3.23.0 / 4.15.0-123-generic, LLVM 6.0.0) for graphics Open GL rendering. In addition, the system consists of two SSDs and one hard disk of 500GB each for data saving. The system uses ROS \cite{ros} Noetic for recording data from the sensors. 

\subsection{Sensor Placement}

\begin{figure}[h!]
    \centering
    \centering
    \includegraphics [scale = 0.7] {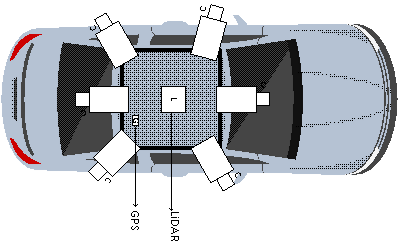}
    \caption{Sensor placement of the instrumented vehicle.}
    \label{fig:SesnorPlcaement}
\end{figure}

The main goal of the vehicle-based data acquisitions system is to cover 360 degrees around the vehicle and record accurate GPS information. Four cameras are placed at the corners on the top of the vehicle, and two cameras are placed at the center, one front-facing and one back facing. The cameras on corners are equipped with wide-angle lenses, and the cameras at the center are equipped with 40-degree lenses. The LiDAR is placed at the center of the vehicle's roof to provide maximum coverage around the vehicle. The GPS is put at the back of the vehicle and on the roof.  Fig.~\ref{fig:SesnorPlcaement} illustrates the placement of sensors on the instrumented vehicle. 

\section{Overview of the Dataset}

The dataset is developed predominantly using the data from GPS and LiDAR, with the camera images as a reference. Each scenario is sixty seconds in length and consists of information about the ego vehicle (waypoints, velocity, and yaw angle), non-ego actors (waypoints, velocity, yaw angle, entry time, and exit time), and road information (name, centers, and banking angle). The dataset provides driving scenarios \cite{Park2020CreatingDS} with a diversity in driving behaviors.   
\subsection{GPS Processing}

\begin{figure}[h!]
    \centering
    \centering
    \includegraphics [scale = 0.45] {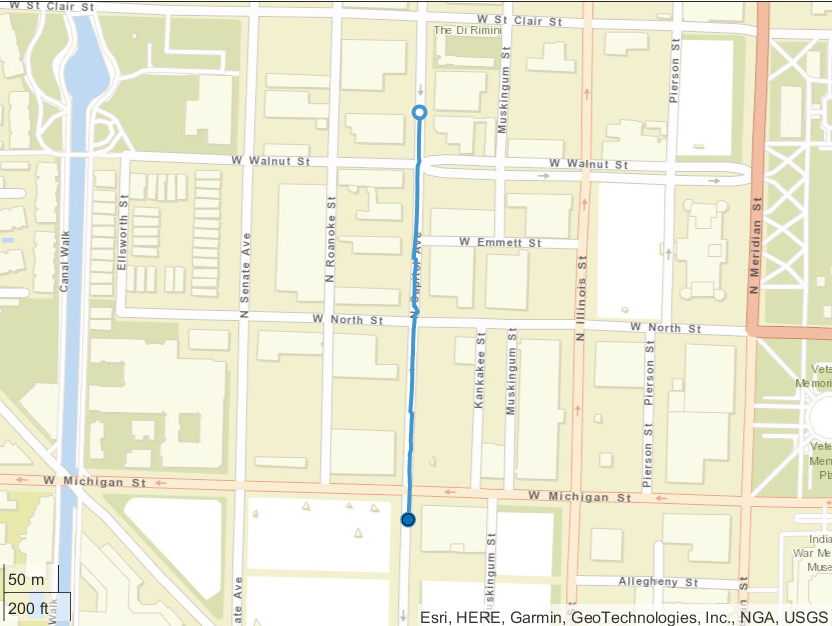}
    \caption{Example GPS data.}
    \label{fig:GPSdata}
\end{figure}

The GPS data is recorded as bag files, then extracted to obtain latitude, longitude, elevation, yaw rate, velocity, and timestamp. The map area covered by the GPS is then downloaded from the OpenStreetMap \cite{OpenStreetMap}. The road network is then created using MATLAB scenario generator along with road attributes. In this paper, the lane information is not included. It will be incorporated in the future using high-definition map service. The GPS data is filtered using Savitzkey - Golay filter \cite{SGolayF}, \cite{Park2020CreatingDS} and is used to compute the local trajectory and velocity of the ego vehicle. Fig.~\ref{fig:GPSdata} shows the example GPS data collected in a driving scenario. 

\subsection{LiDAR Processing}

    \subsubsection{Data Collected}
\begin{figure}[h!]
        \centering
        \centering
        \includegraphics [scale = 0.5] {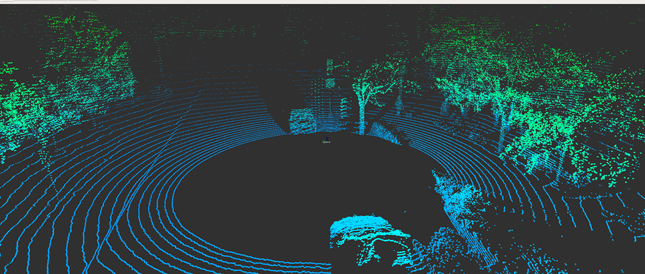}
        \caption{Example LiDAR data.}
        \label{fig:Lidardata}
    \end{figure}

    The LiDAR data is recorded as UDP packets for better transfer speed and lossless recording. The UDP packets are also stored in bag files. These bag files are then extracted and the UDP packets are converted to MATLAB point-cloud object with their respective timestamps. The relative time of each point-cloud frame with respect to the GPS timestamp is obtained. The LiDAR data is then used to detect and track the non-ego actors. Fig. \ref{fig:Lidardata} shows exemplary point-cloud data recorded from LiDAR. 
    
    \subsubsection{Object Detection}
    A virtual driving scenario was created with the GPS, IMU, and LiDAR data. The GPS data was used to show the trajectory and movement of the targets on a high-definition map. These scenarios allow the simulation and testing of different motion planning algorithms. 
 
    The Non-Ego actors were detected using the Joint Probabilistic Data Association algorithm\cite{SonarJDPA} implemented on MATLAB. It allows for the detection of multiple targets. This algorithm enables estimation of states of unknown and time-varying targets from noisy and uncertain environments, which measures and calculates the hypothesis pairs for every agent for every track. Each track measurement hypothesis has a weight, which is calculated by the normalized sum of the conditional probabilities of the agent track association\cite{PDA}. The tracker employs joint probabilistic data association to assign detection to each track. Multiple detection can contribute to each track; hence the tracker uses a soft assignment. Tracks are created, confirmed, corrected, predicted, and deleted by the tracker. In MATLAB, Detection reports created by \emph{objectDetection} object are used as inputs to the tracker\cite{MatlabJDPA}. 
     
   Any new track begins with a tentative state. If a tentative track receives enough detection, its status is changed to confirmed. The related track is instantly confirmed if the detection has a known classification. The tracker considers a track as a physical object once it has been confirmed. The track is removed if no detection is assigned within a certain amount of updates. The tracker calculates the state vector and state estimate error covariance matrix for each track\cite{MatlabJDPA}. Each detection has at least one track associated with it. If the detection cannot be assigned to an existing track, the tracker creates a new one. If a tracker maintains multiple tracks, the data association process becomes more complicated because one detection can fall within the validation gates of multiple tracks. In our case, clustering is used since there are multiple actors and tracks in one scenario.
   
   The position and velocity of the surrounding vehicles in relation to the ego vehicle are provided by LiDAR detection. We converted the target vehicle positions to global coordinates and approximated the orientation angle based on vehicle motion using Equation (\ref{eqn:ObjectDetection}) below. 
   
   \begin{equation}
       (X_t, Y_t) = (X_{ego}, Y_{ego}) + R(\psi_{ego})\cdot(x_t, y_t)^T,
       \label{eqn:ObjectDetection}
   \end{equation}
   where $R$ is the Rotation Matrix, $(x_t, y_t)$ is the position in the ego vehicle's coordinate, and $(X_t, Y_t)$ is the position in the global coordinate. 
   
In scenarios where there are multiple non-ego actors close to each other, an issue arises where the tracks for all vehicles are noisy, and a track for a vehicle changes from one lane to another. This issue is resolved by reducing the default track difference length in MATLAB, which allows separate tracks for the respective non-ego actors. However, when the track difference length is small, noisy data can be included in the scenario. These noisy tracks are manually deleted by verifying the non-ego actors with the camera.  

Since the Joint Probabilistic Data Association algorithm performs a soft assignment, it occasionally allows two non-ego actors very close to each other to be considered as one non-ego actor. This is again fixed by manually verifying it with the camera and changing the track assignment size so that each non-ego actor gets its own track.
   
\subsection{Camera Data}
\begin{figure}[h]
    \centering
    \centering
    \includegraphics [scale = 0.17] {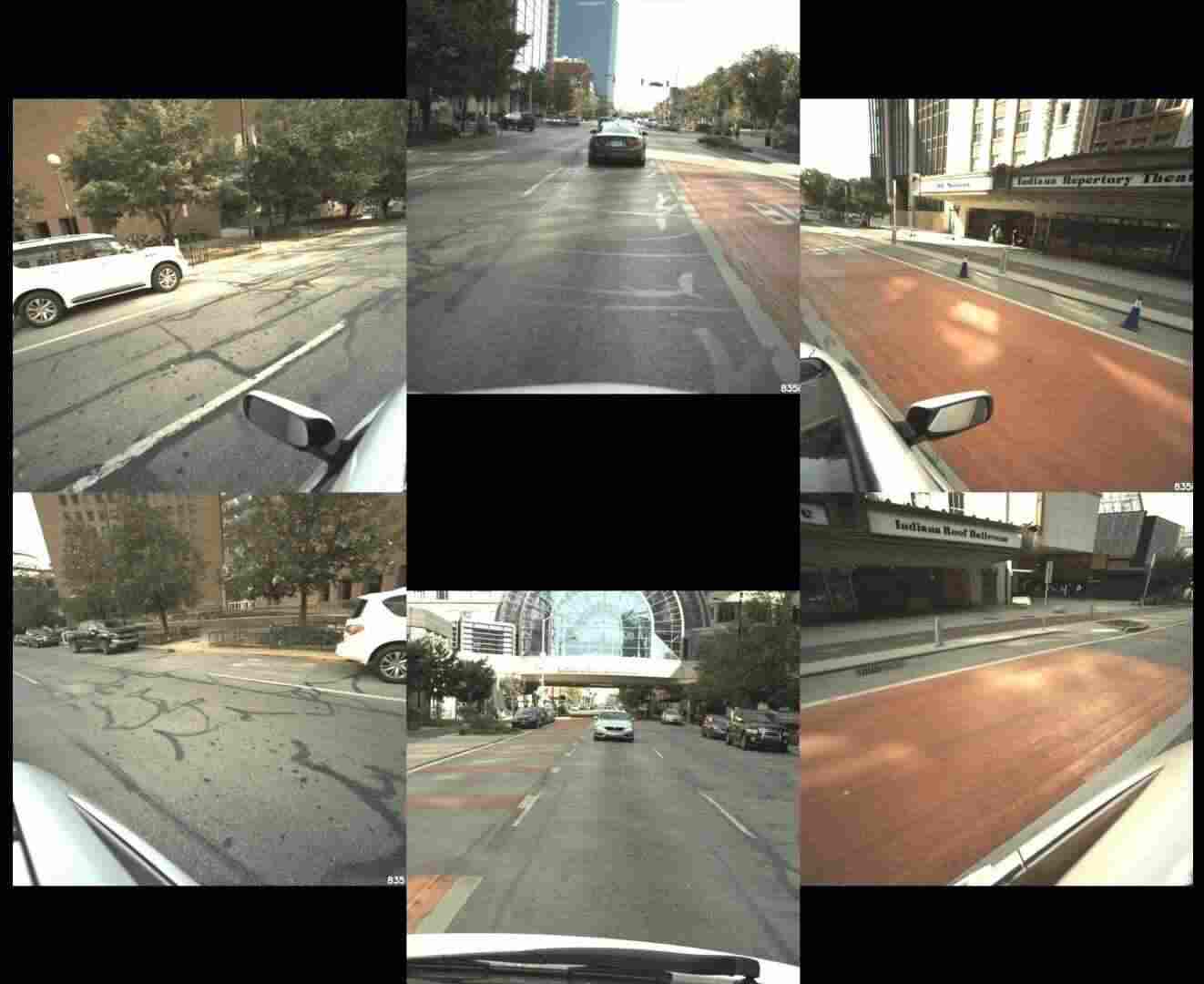}
    \caption{Example Camera data.}
    \label{fig:Cameradata}
\end{figure}
Even though the camera data is currently used as reference for the generated scenarios, it will be used to add much more information to the dataset in the future. An example camera view from all six cameras is shown in Fig. \ref{fig:Cameradata}.

\section{Scenario Development from the Dataset}
\begin{figure}[h!]
    \centering
\begin{subfigure}{0.3\linewidth}
\includegraphics [width=\linewidth]{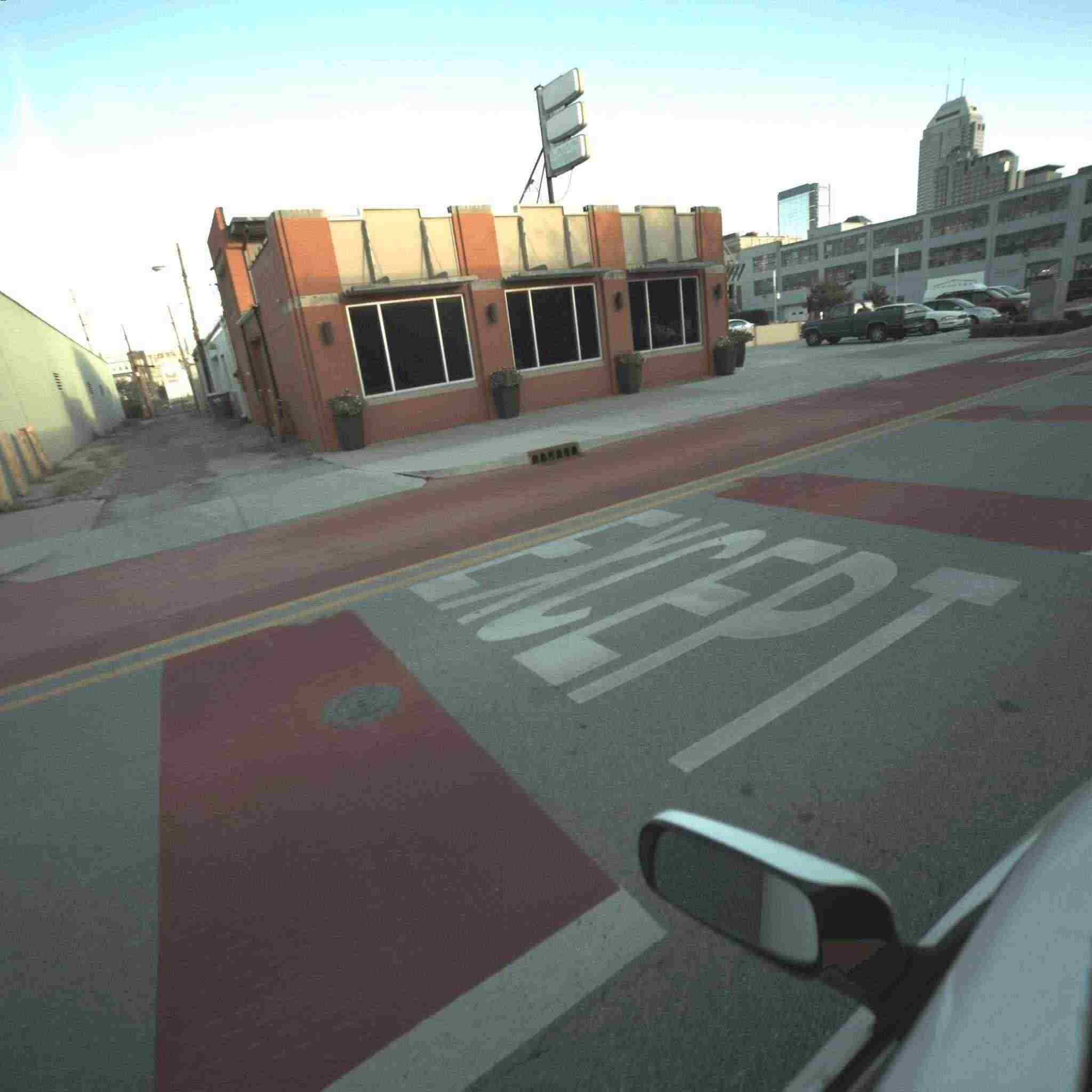}
\caption{Left Front}
\label{fig:cam_left_front}
\end{subfigure}
~
\begin{subfigure}{0.3\linewidth}
\includegraphics [width=\linewidth]{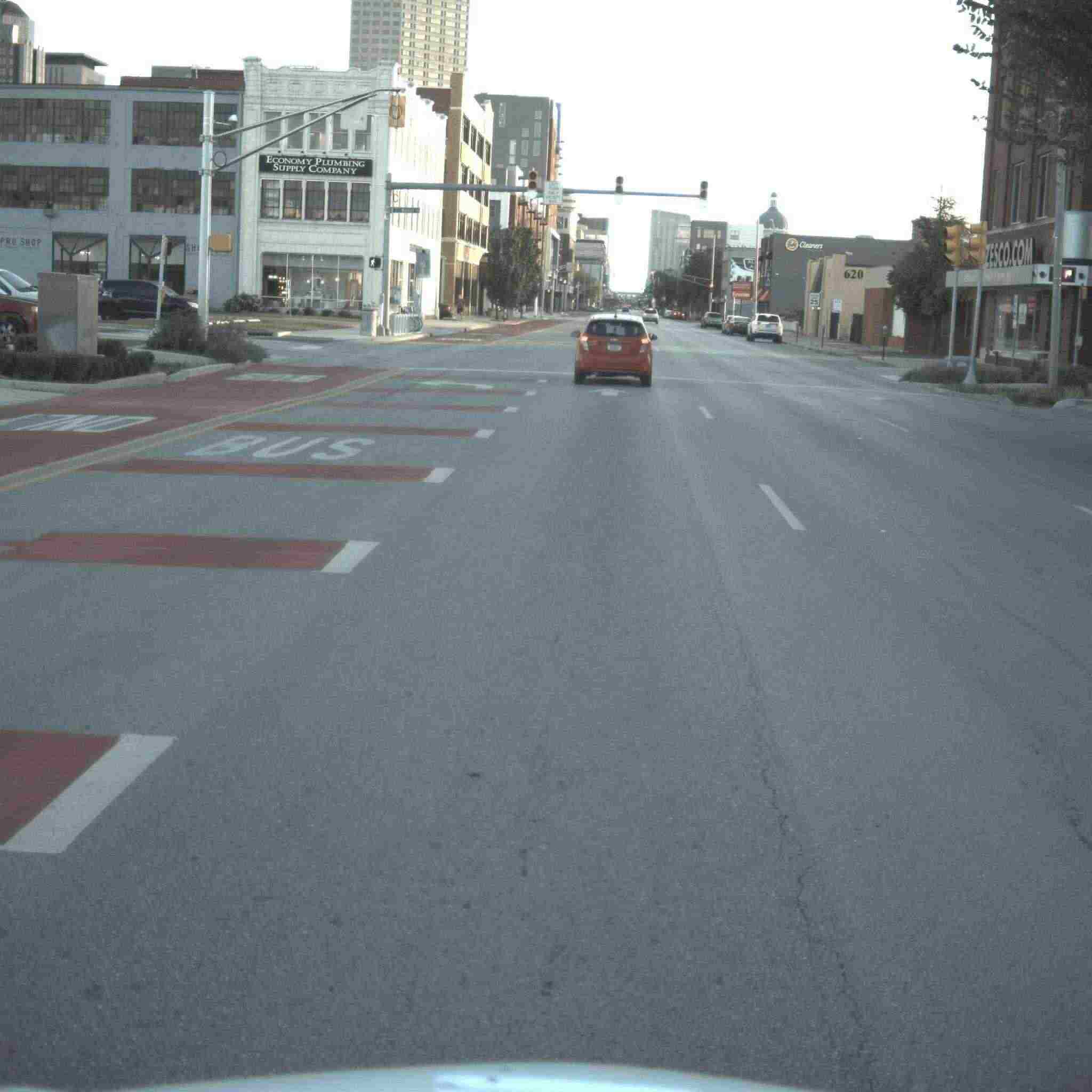}
\caption{Front Center}
\label{fig:cam_front}
\end{subfigure}
~
\begin{subfigure}{0.3\linewidth}
\includegraphics [width=\linewidth]{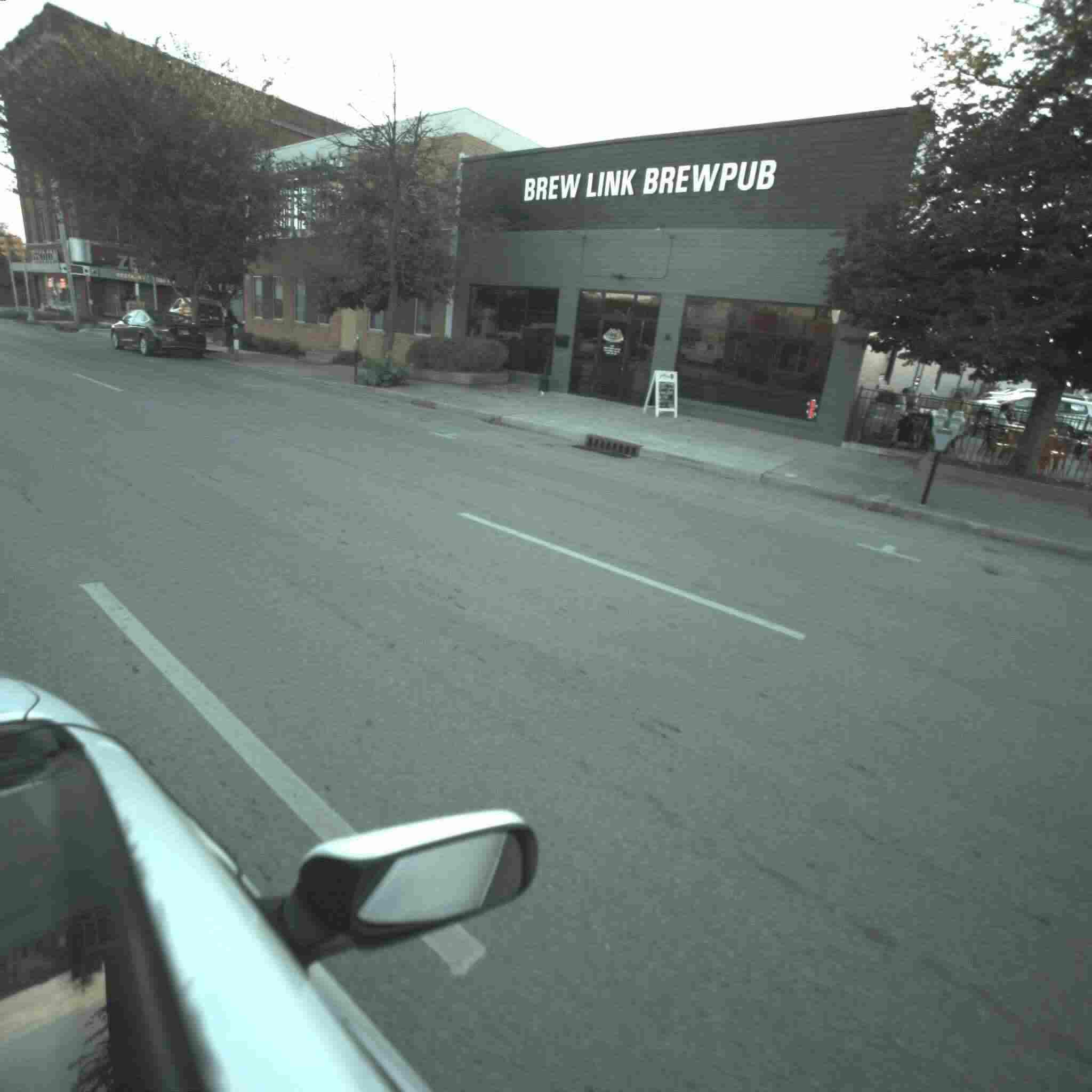}
\caption{Right Front}
\label{fig:cam_right_front}
\end{subfigure}
~
\begin{subfigure}{0.3\linewidth}
\includegraphics [width=\linewidth]{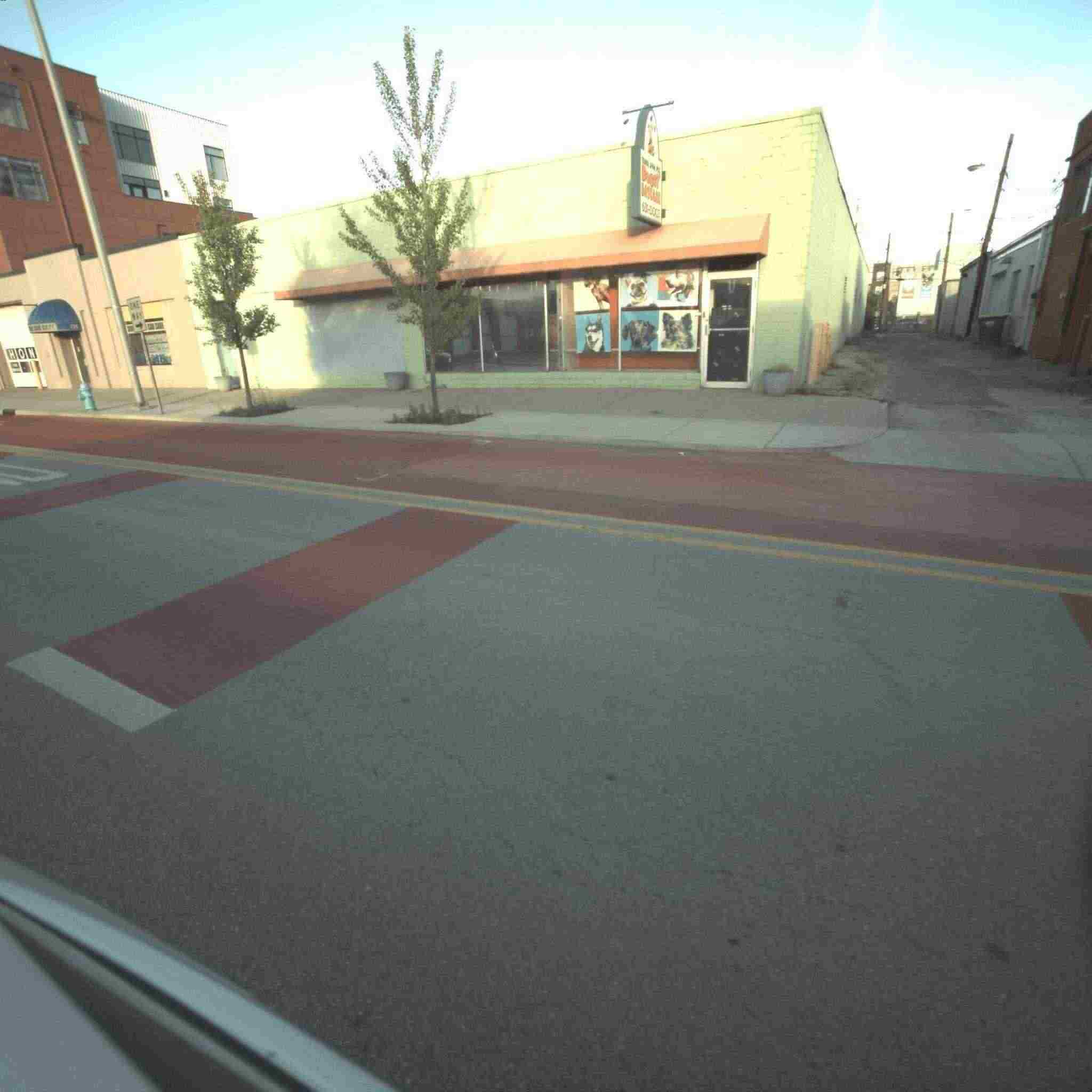}
\caption{Left Back}
\label{fig:cam_left_back}
\end{subfigure}
~
\begin{subfigure}{0.3\linewidth}
\includegraphics [width=\linewidth]{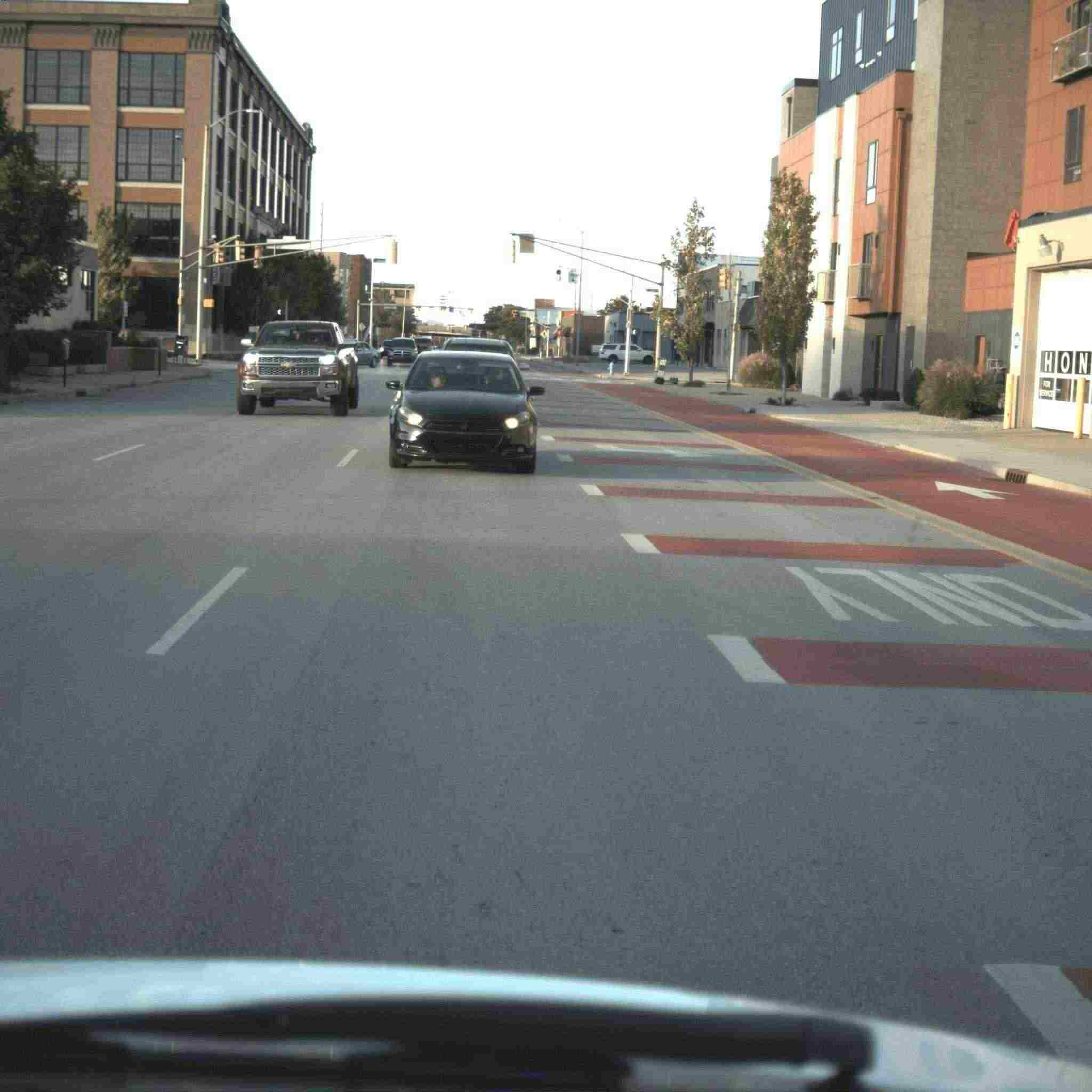}
\caption{Back Center}
\label{fig:cam_back}
\end{subfigure}
~
\begin{subfigure}{0.3\linewidth}
\includegraphics [width=\linewidth]{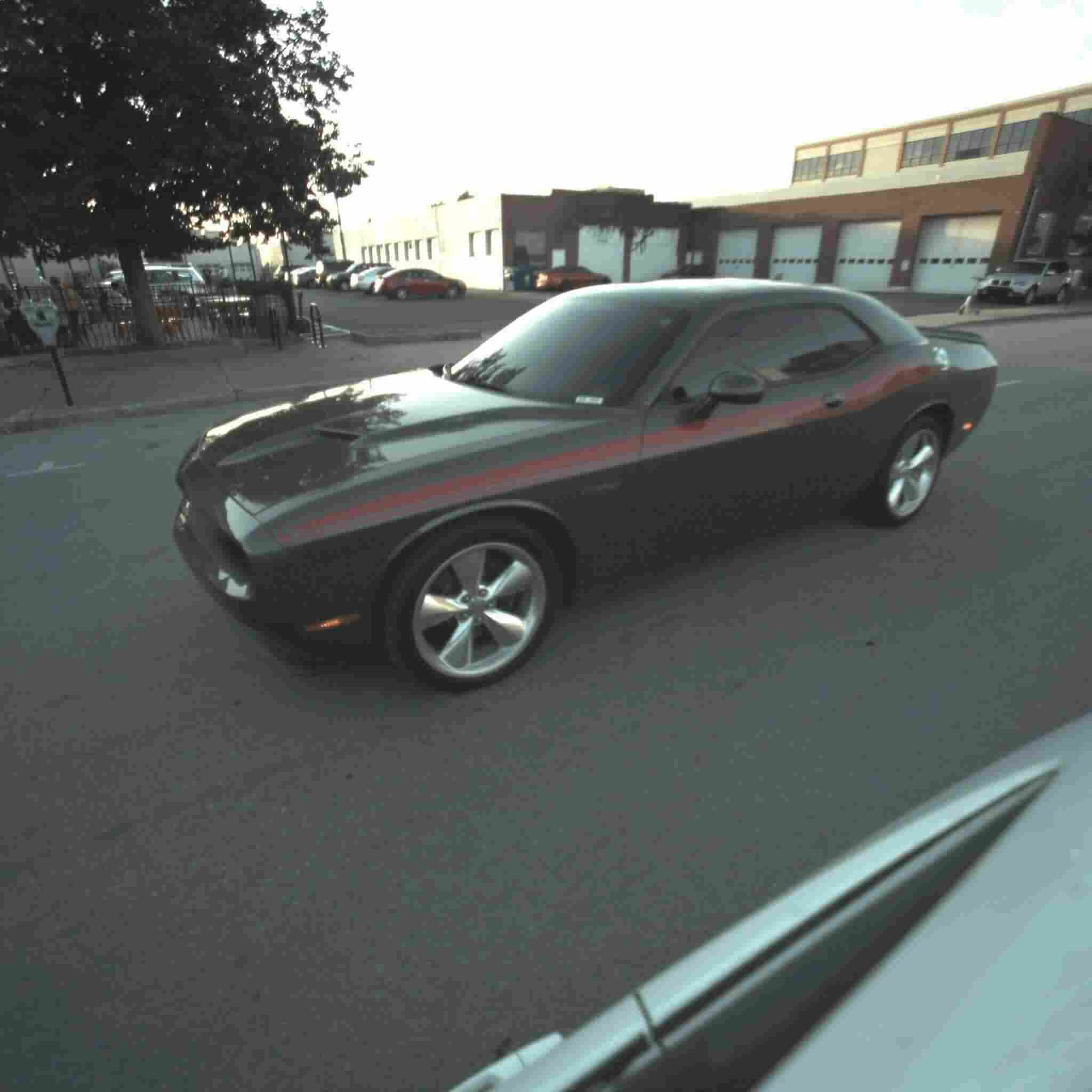}
\caption{Right Back}
\label{fig:cam_right_back}
\end{subfigure}
\caption{Reference camera frames - Scenario 1.}
\label{fig:Cam_Ref_Scenario 1}
\end{figure}

\begin{figure}[h!]
    \centering
\begin{subfigure}{0.7\linewidth}
\includegraphics [width=\linewidth]{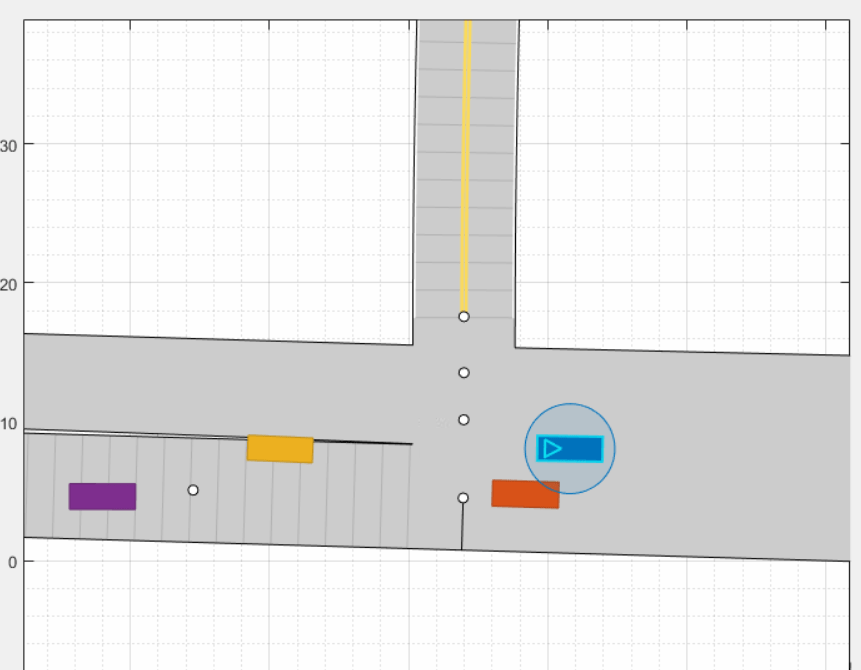}
\caption{Bird's-eye view.}
\label{fig:Birdeye-1}
\end{subfigure}
~
\begin{subfigure}{0.7\linewidth}
\includegraphics [width=\linewidth]{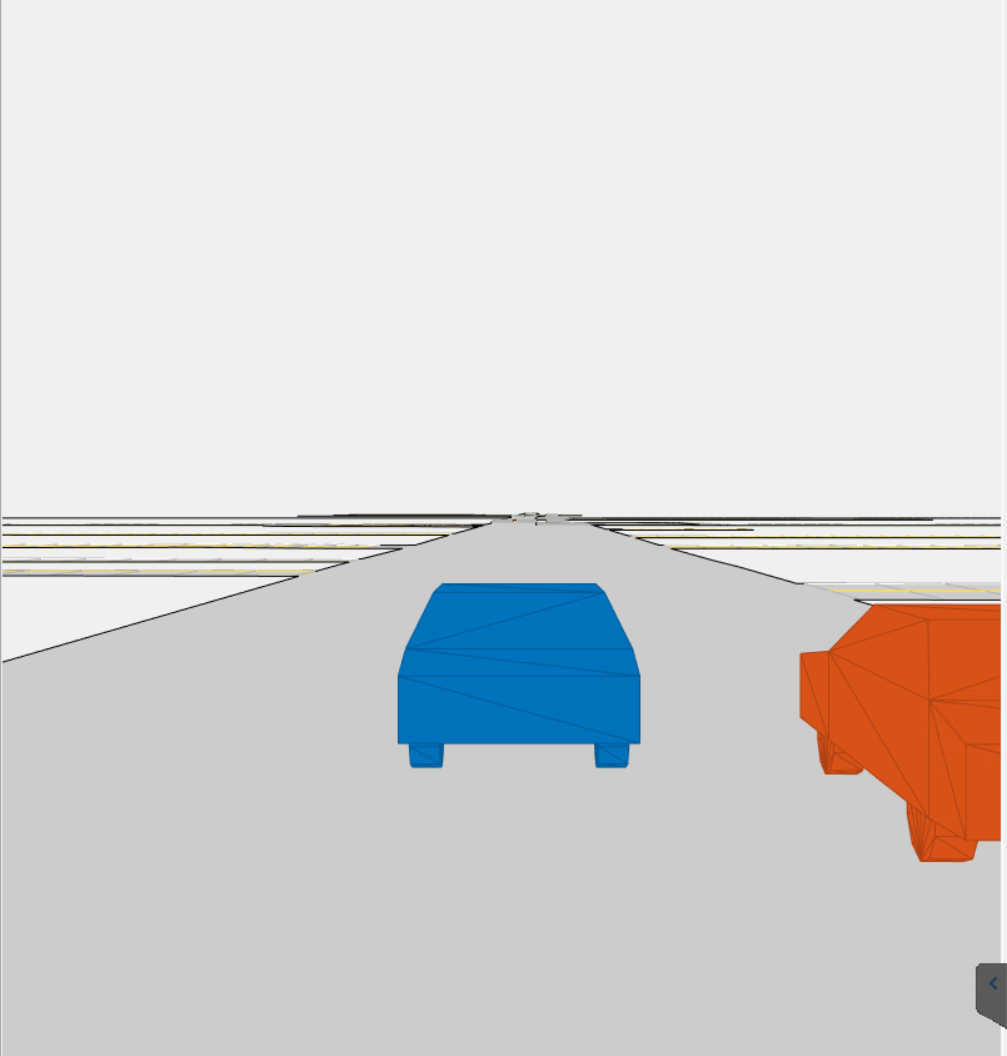}
\caption{Ego-centric view.}
\label{fig:egocentric-1}
\end{subfigure}
\caption{
Generated Scenario 1.}
\label{fig:Scenario_1_screenshots}
\end{figure}

 This section describes several scenario development as examples and with reference images. Scenario 1 describes a case where the ego vehicle is in interaction with two other vehicles, one at the back and one on the right side in the blind-spot range. This is depicted in the camera images in Fig. \ref{fig:Cam_Ref_Scenario 1}. The equivalent scenario generated using Matlab is shown in Fig. \ref{fig:Scenario_1_screenshots}. The ego-vehicle is depicted and encircled in blue. Comparing the images in Fig. 9(e) and Fig. 9(f), it can be seen that it matches with the three vehicles in the generated scenario. It can also be seen that the vehicle captured by the front center camera, Fig. 9(b), is not generated in the scenario. This is because the distance between the ego vehicle and the vehicle in front is large. This issue can be addressed where  cameras are also involved in actor detection. 
 
\begin{figure}[h!]
    \centering
\begin{subfigure}{0.3\linewidth}
\includegraphics [width=\linewidth]{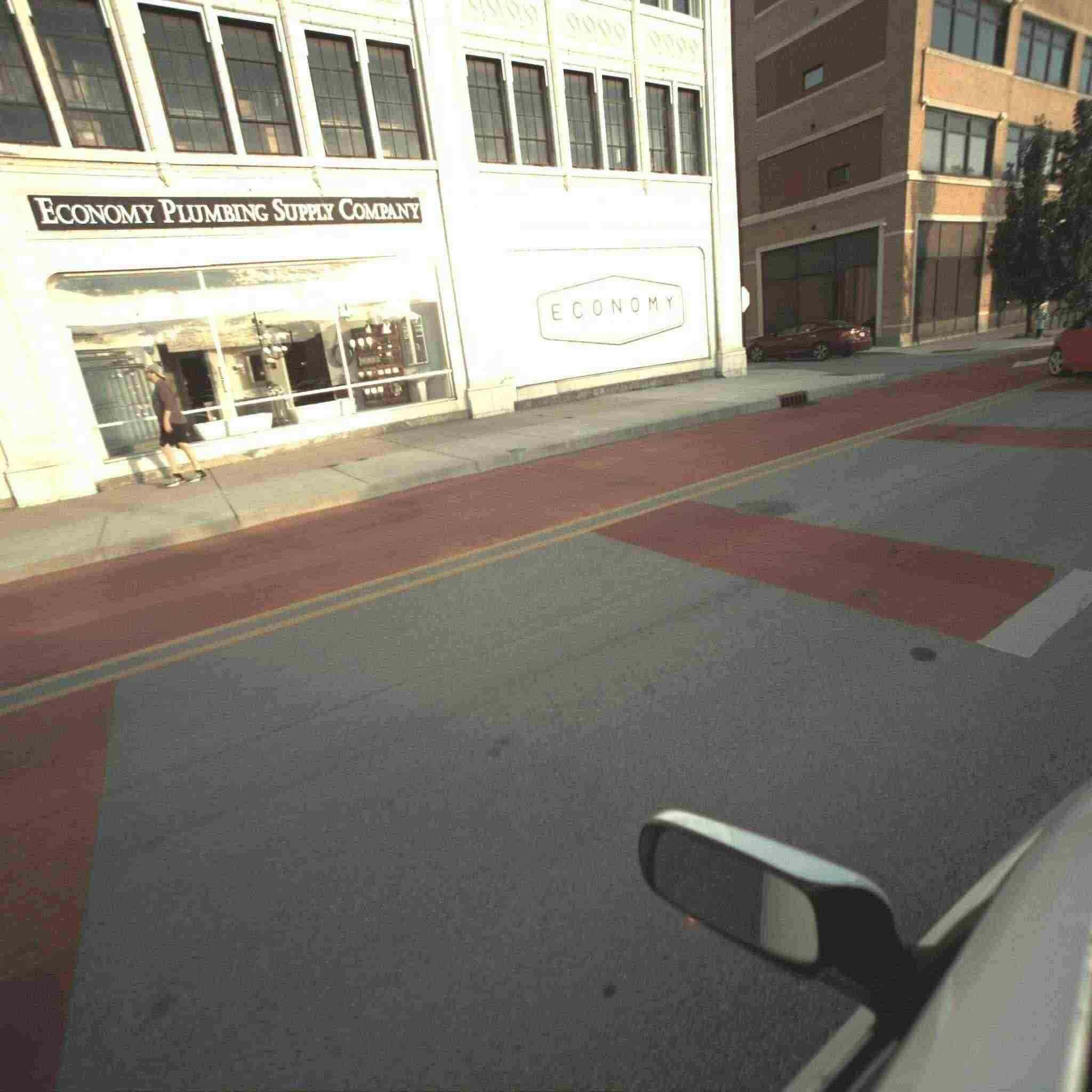}
\caption{Left Front}
\label{fig:cam_left_front}
\end{subfigure}
~
\begin{subfigure}{0.3\linewidth}
\includegraphics [width=\linewidth]{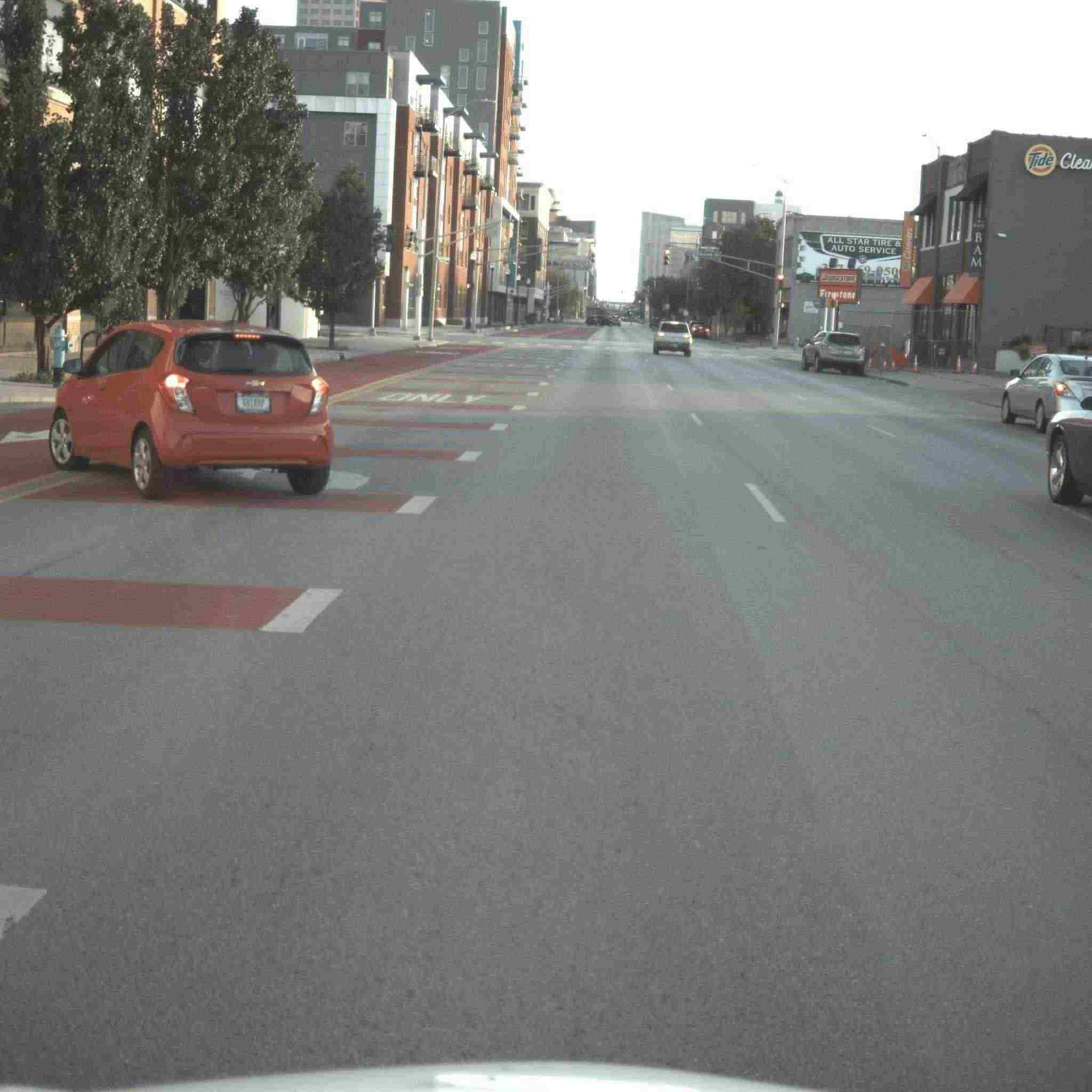}
\caption{Front Center}
\label{fig:cam_front}
\end{subfigure}
~
\begin{subfigure}{0.3\linewidth}
\includegraphics [width=\linewidth]{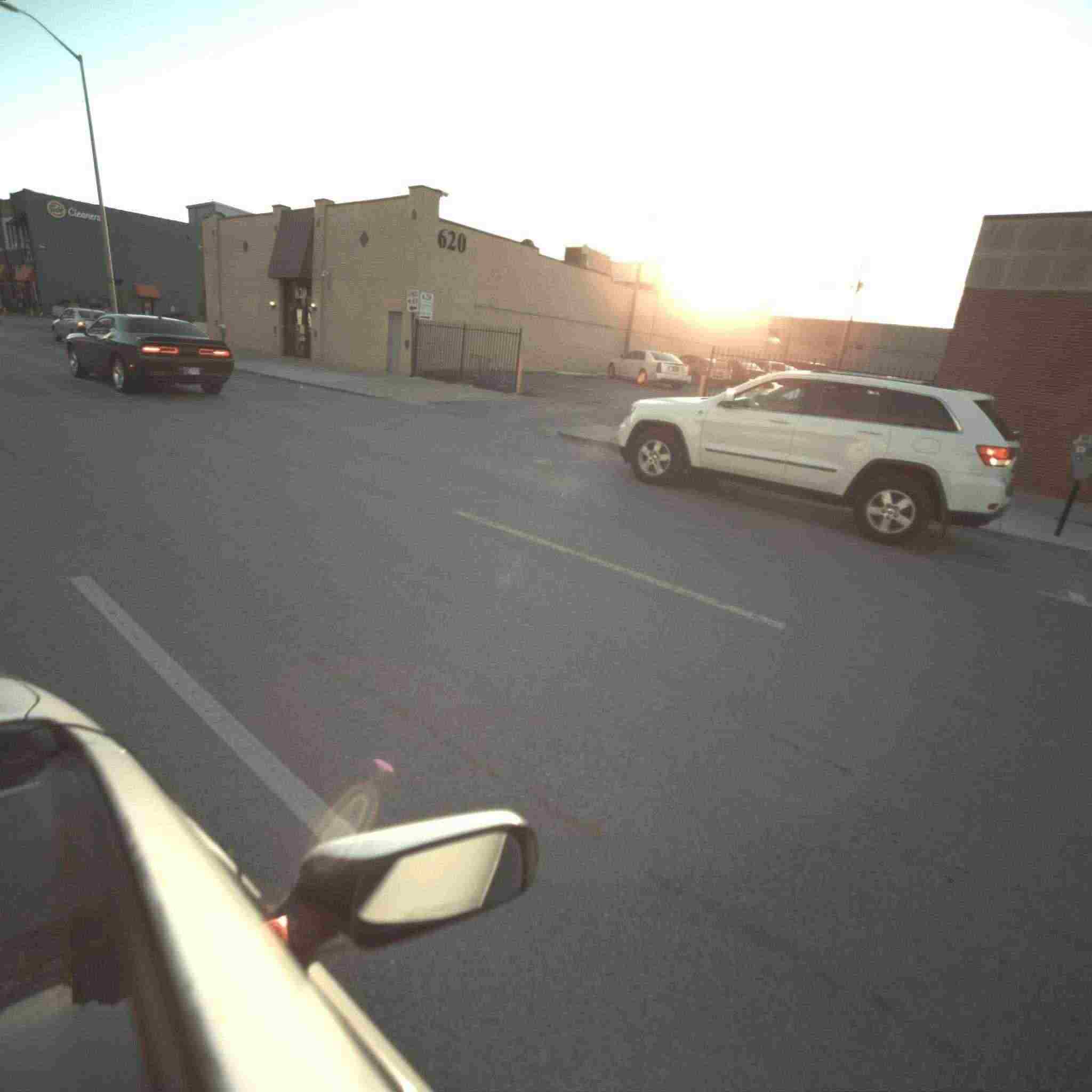}
\caption{Right Front}
\label{fig:cam_right_front}
\end{subfigure}
~
\begin{subfigure}{0.3\linewidth}
\includegraphics [width=\linewidth]{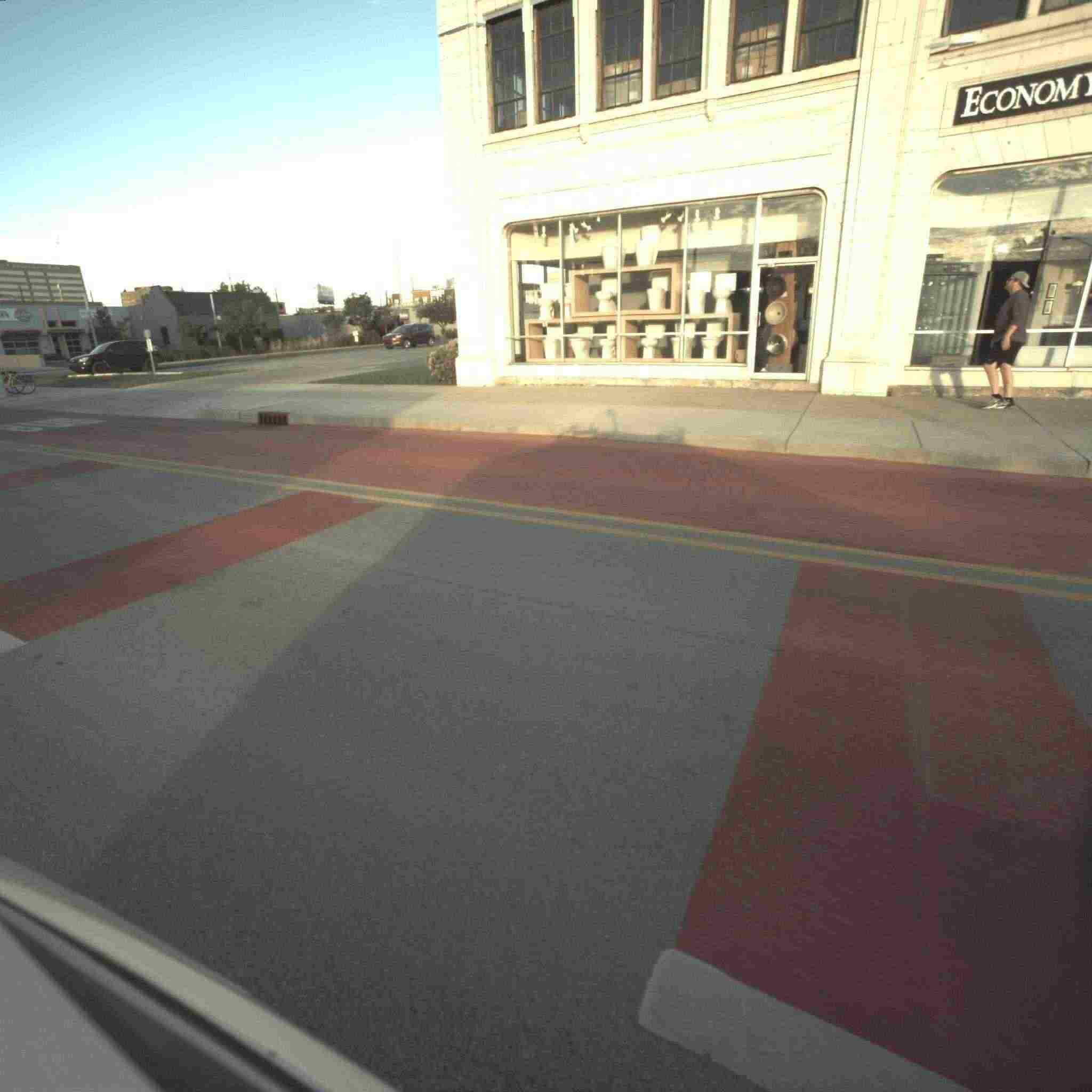}
\caption{Left Back}
\label{fig:cam_left_back}
\end{subfigure}
~
\begin{subfigure}{0.3\linewidth}
\includegraphics [width=\linewidth]{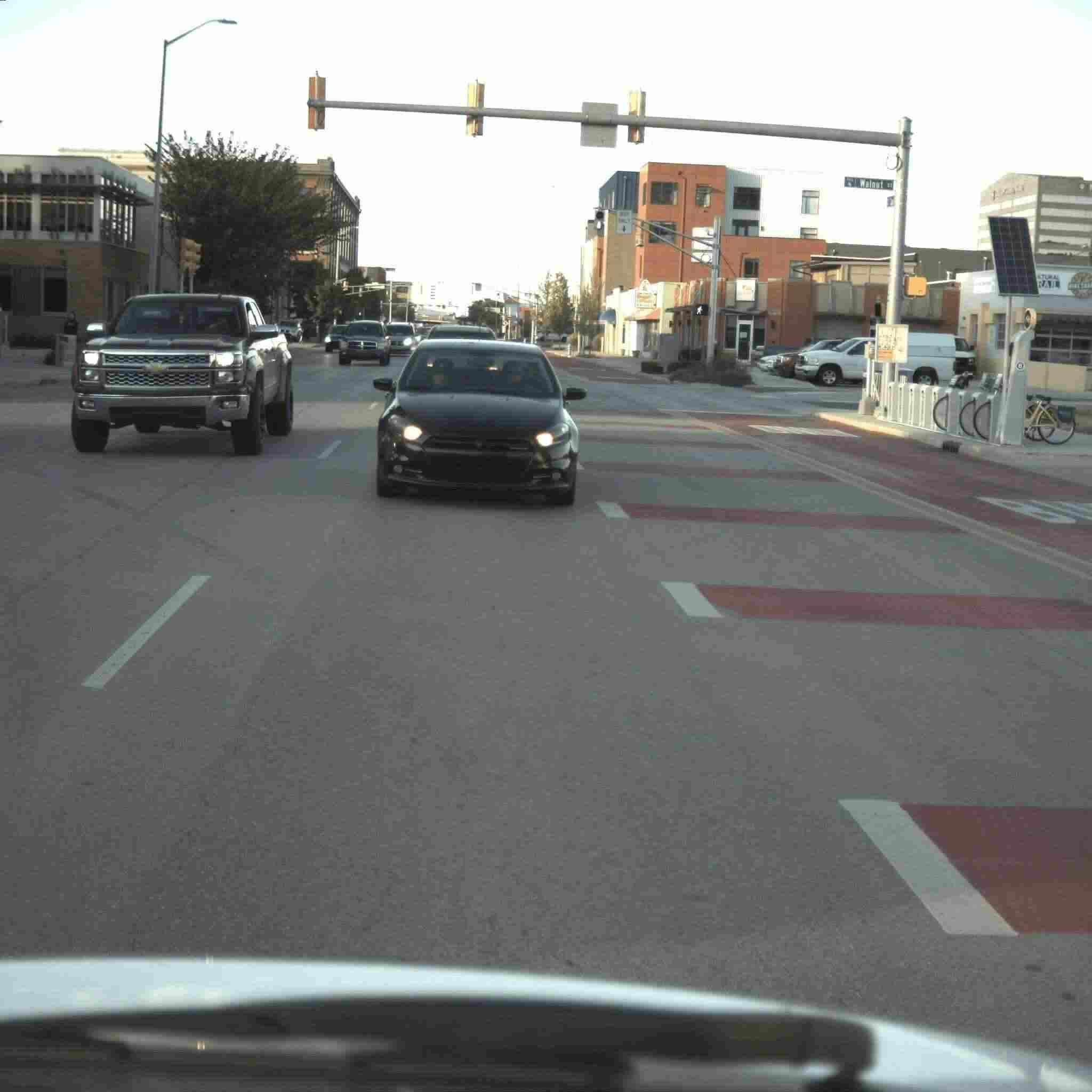}
\caption{Back Center}
\label{fig:cam_back}
\end{subfigure}
~
\begin{subfigure}{0.3\linewidth}
\includegraphics [width=\linewidth]{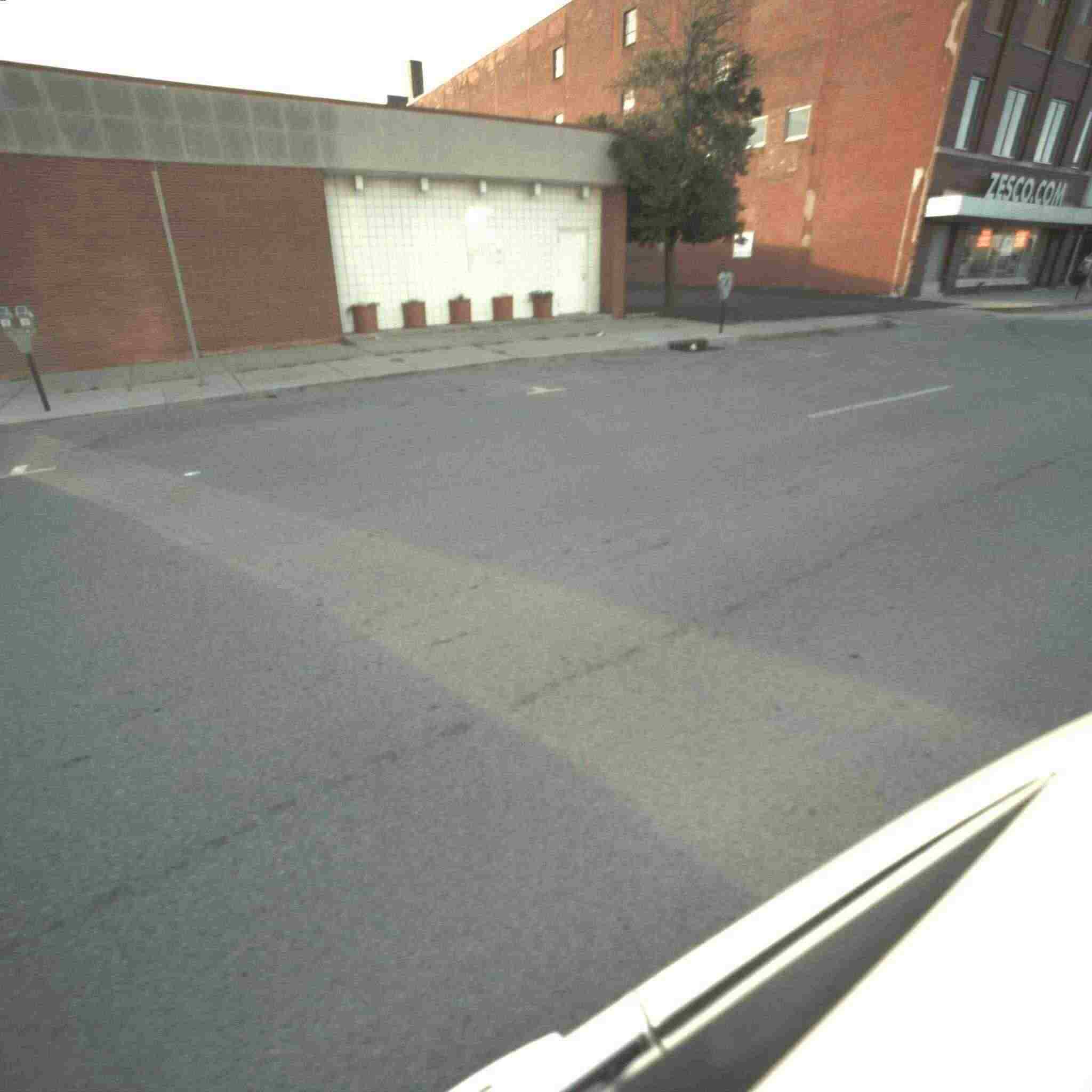}
\caption{Right Back}
\label{fig:cam_right_back}
\end{subfigure}
\caption{Reference camera frames - Scenario 2.}
\label{fig:Cam_Ref_Scenario 2}
\end{figure}

\begin{figure}[h]
    \centering
\begin{subfigure}{0.7\linewidth}
\includegraphics [width=\linewidth]{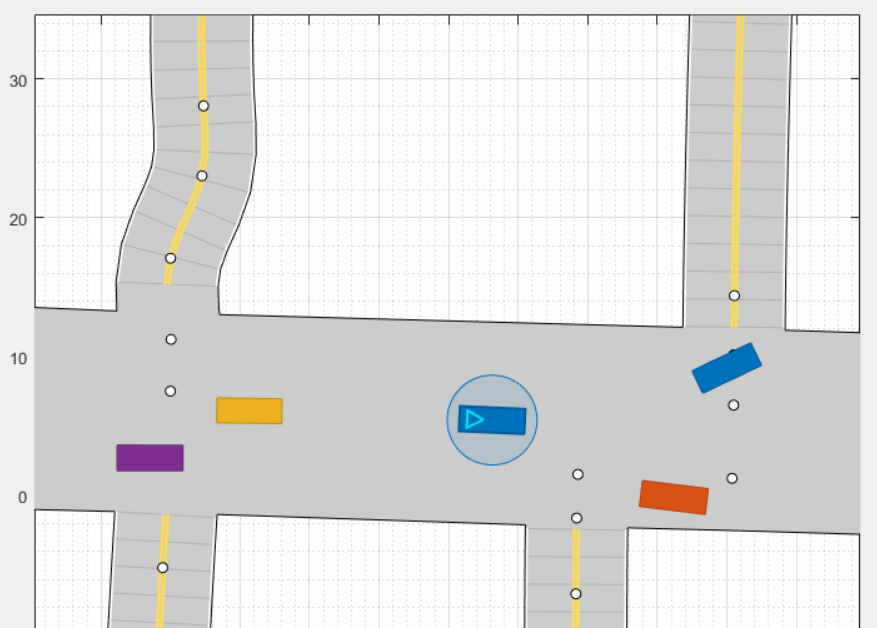}
\caption{Bird's-eye view.}
\label{fig:Birdeye-2}
\end{subfigure}
~
\begin{subfigure}{0.7\linewidth}
\includegraphics [width=\linewidth]{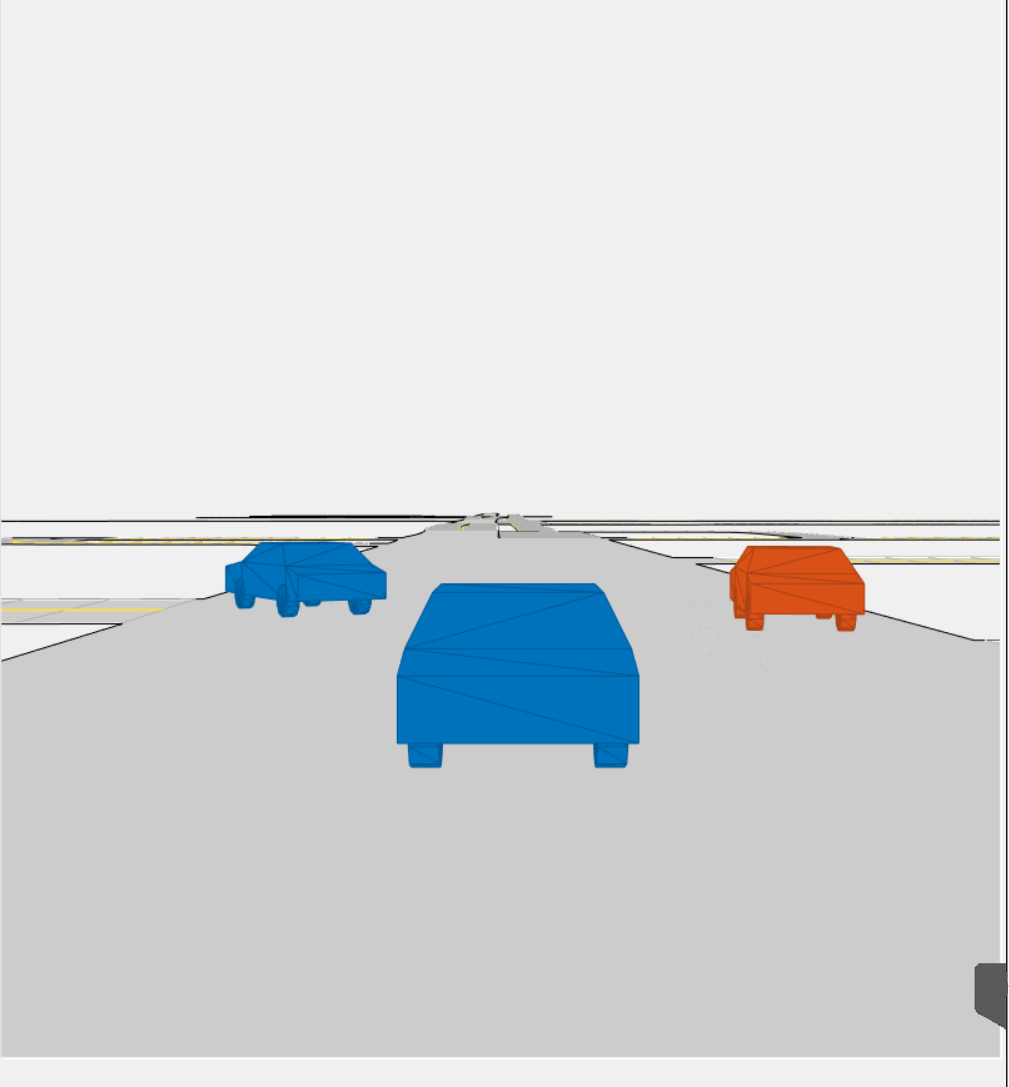}
\caption{Ego-centric view.}
\label{fig:egocentric-2}
\end{subfigure}
\caption{
Generated Scenario 2.}
\label{fig:Scenario_2_screenshots}
\end{figure}

Scenario 2 depicts the situation where the vehicle in front slows down, changes lane to the left, and turns left. This can be seen in the reference images in Fig. \ref{fig:Cam_Ref_Scenario 2}. In Fig. \ref{fig:Cam_Ref_Scenario 2}, the front camera, right-front camera, and back camera shows the cars in view of the ego vehicle. It can also be seen in the generated scenario, as illustrated in Fig. \ref{fig:Scenario_2_screenshots}.

\begin{figure}[h!]
    \centering
\begin{subfigure}{0.3\linewidth}
\includegraphics [width=\linewidth]{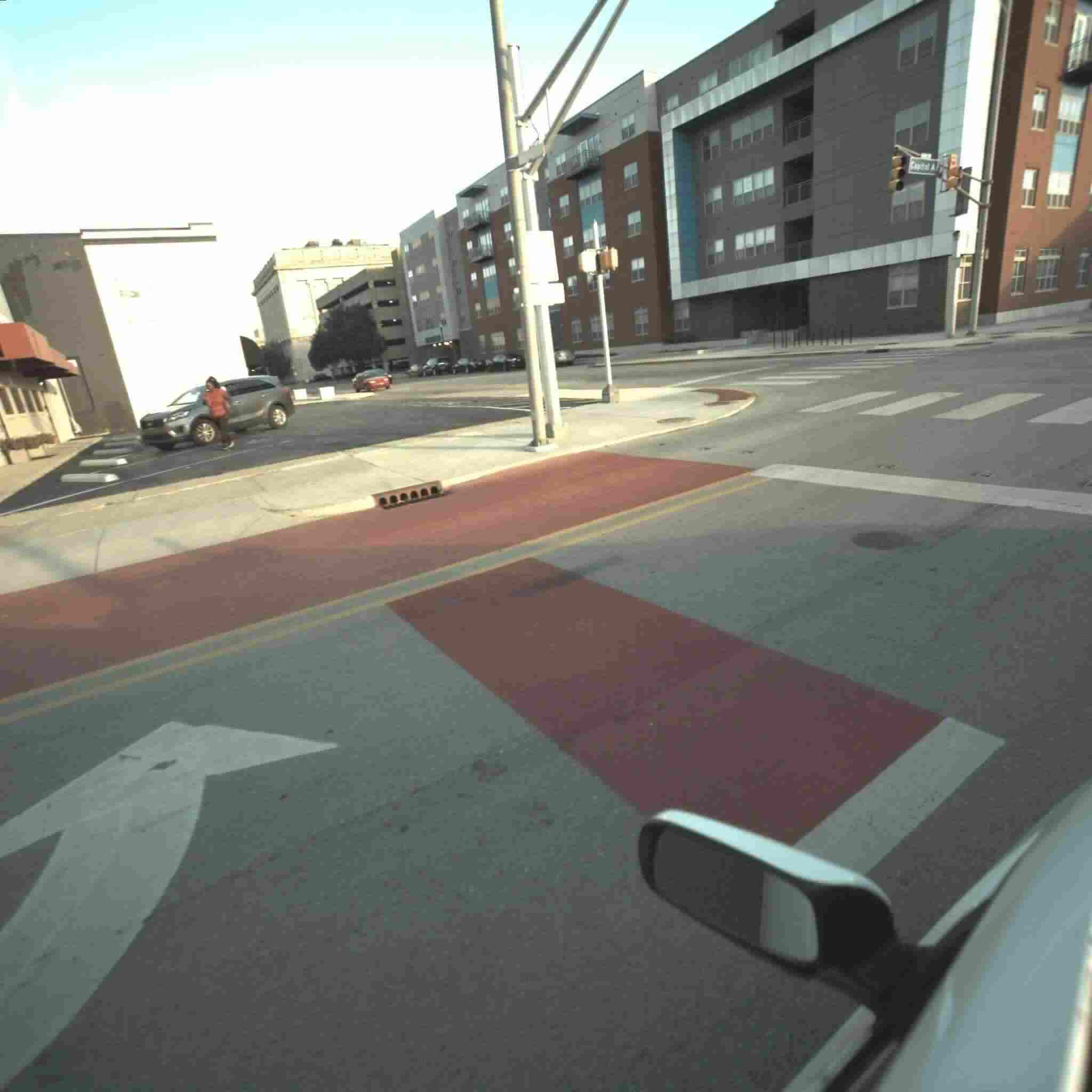}
\caption{Left Front}
\label{fig:cam_left_front}
\end{subfigure}
~
\begin{subfigure}{0.3\linewidth}
\includegraphics [width=\linewidth]{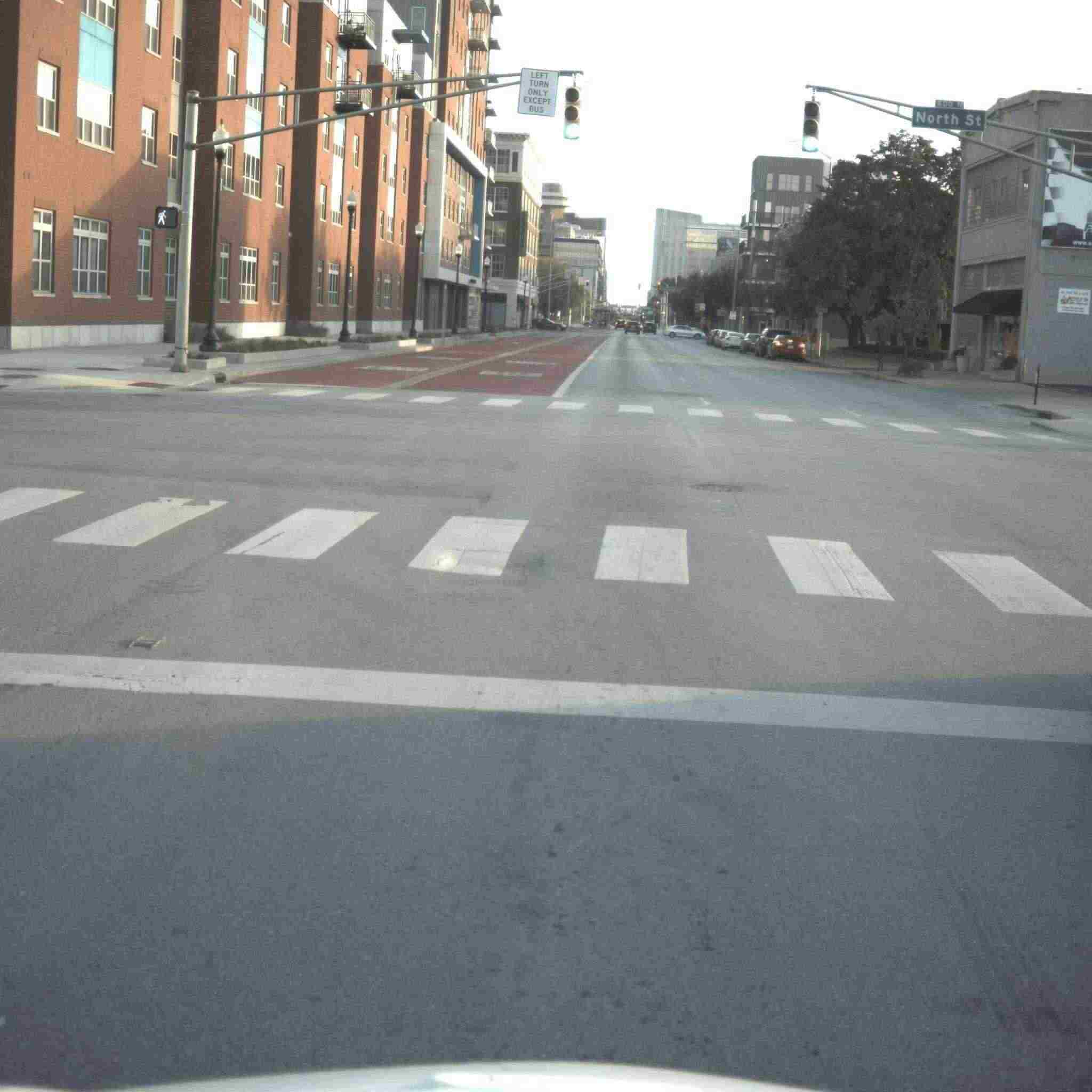}
\caption{Front Center}
\label{fig:cam_front}
\end{subfigure}
~
\begin{subfigure}{0.3\linewidth}
\includegraphics [width=\linewidth]{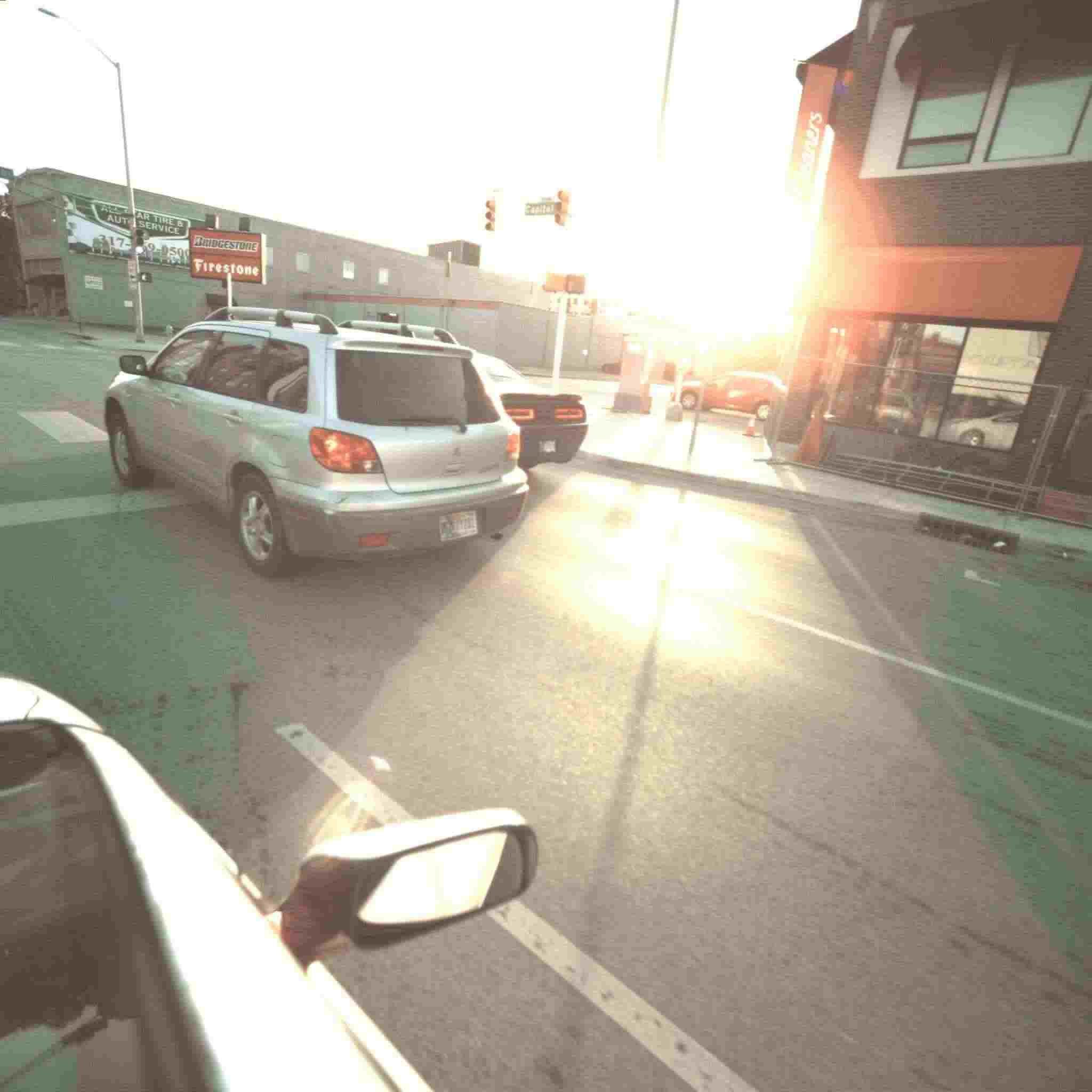}
\caption{Right Front}
\label{fig:cam_right_front}
\end{subfigure}
~
\begin{subfigure}{0.3\linewidth}
\includegraphics [width=\linewidth]{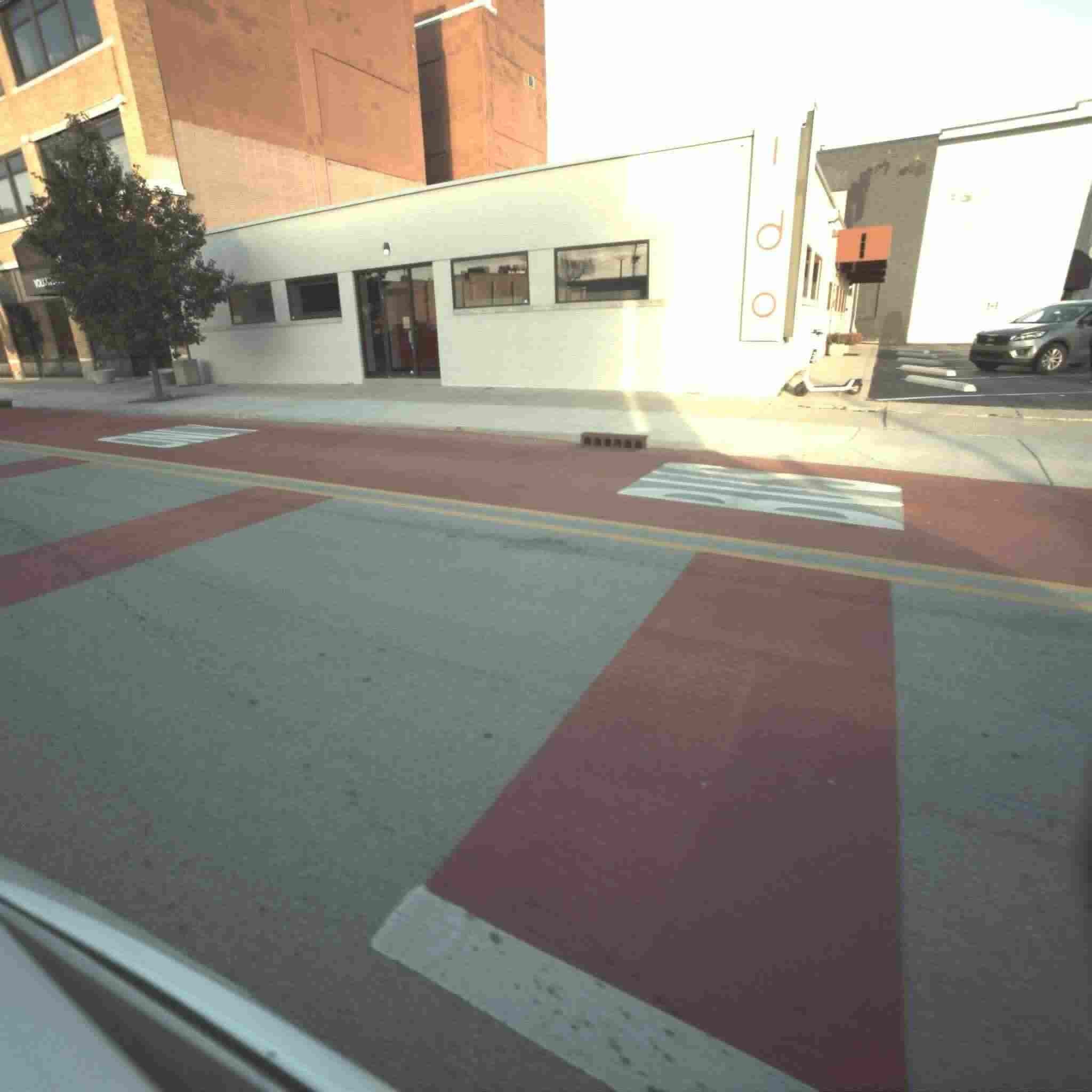}
\caption{Left Back}
\label{fig:cam_left_back}
\end{subfigure}
~
\begin{subfigure}{0.3\linewidth}
\includegraphics [width=\linewidth]{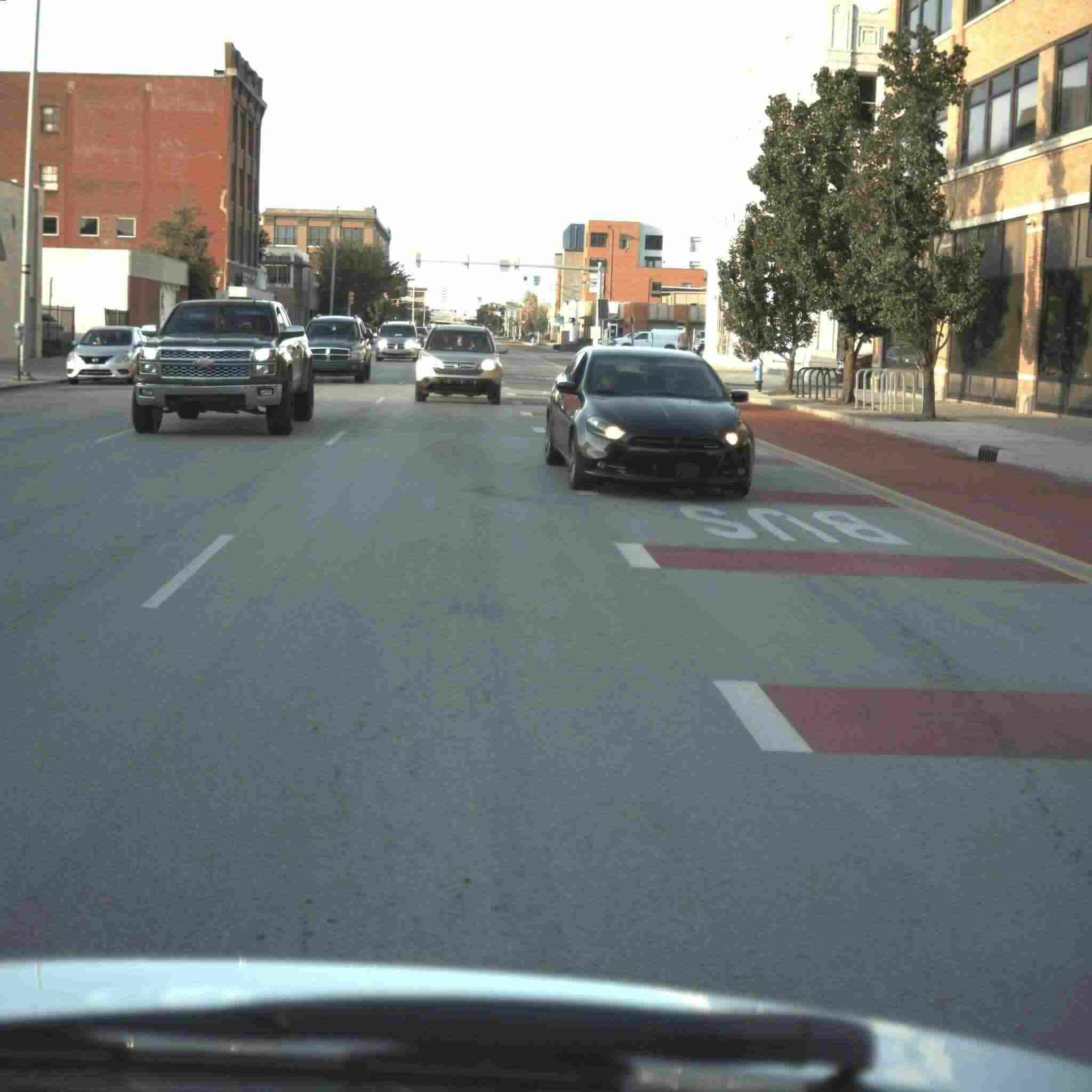}
\caption{Back Center}
\label{fig:cam_back}
\end{subfigure}
~
\begin{subfigure}{0.3\linewidth}
\includegraphics [width=\linewidth]{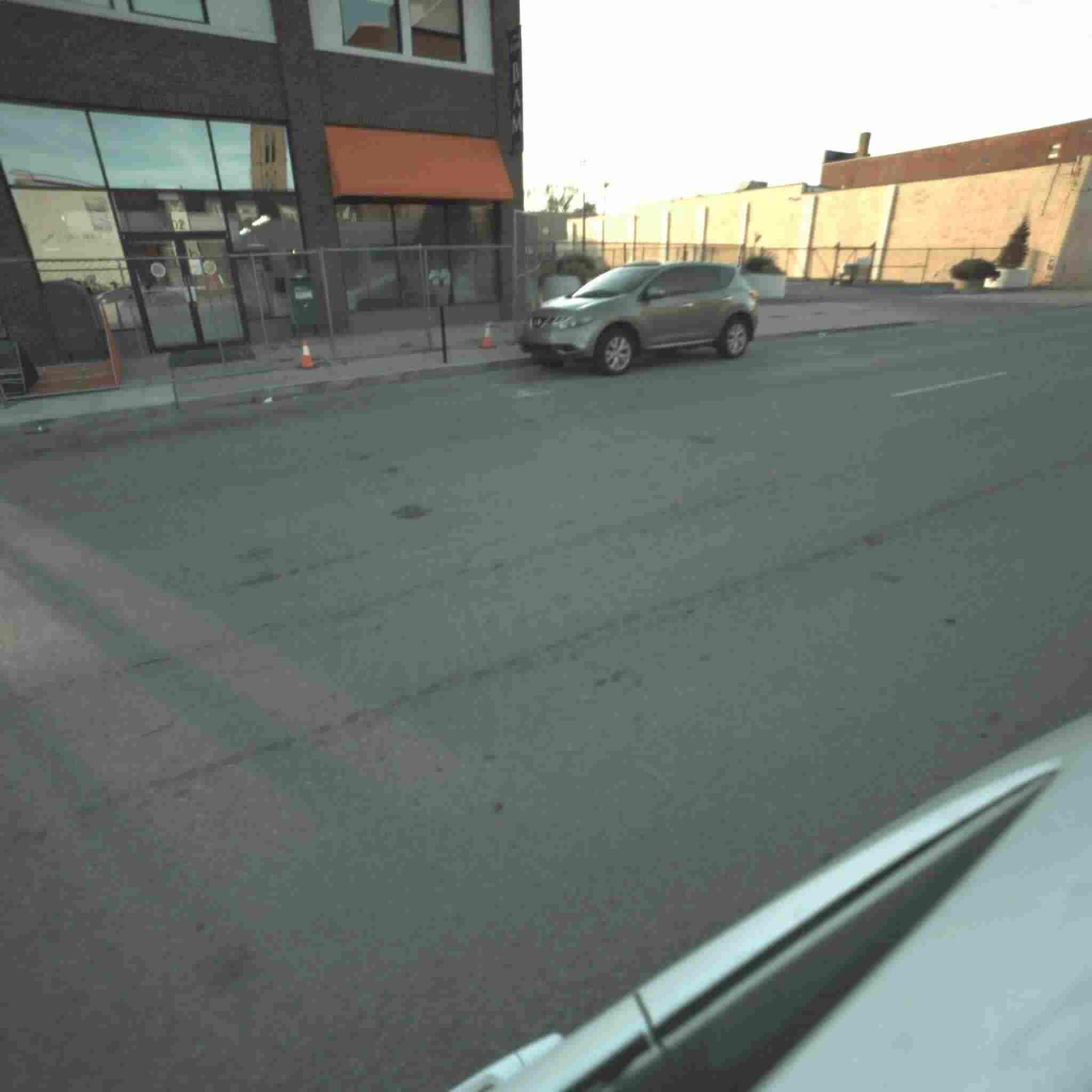}
\caption{Right Back}
\label{fig:cam_right_back}
\end{subfigure}
\caption{Reference camera frames - Scenario 3.}
\label{fig:Cam_Ref_Scenario 3}
\end{figure}


Scenario 3 demonstrates an interaction at a traffic light. The scenario involves a non-ego vehicle which is about to start after a green light. The non-ego vehicle is about to be passed on the left by the ego vehicle and another non-ego vehicle on the right. The camera view is depicted in Fig. \ref{fig:Cam_Ref_Scenario 3}. The generated scenario is not shown due to the page limitation. 

\begin{figure}
    \centering
    \centering
    \includegraphics [scale = 0.5] {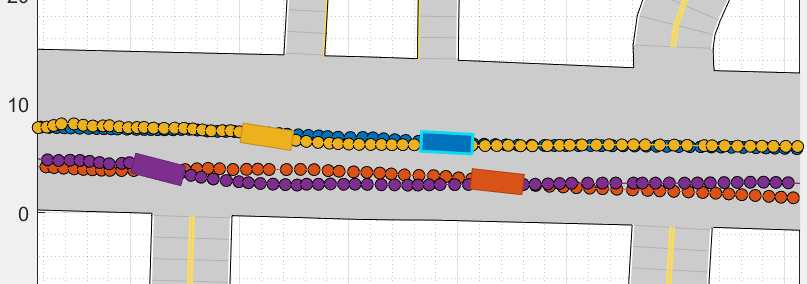}
    \caption{Waypoints of vehicles in a generated scenario.}
    \label{fig:waypoints}
\end{figure}

Note that all scenarios are generated at a frequency of 10~Hz. The scenarios contain information about all ego and non-ego vehicles' waypoints, yaw angle, and the speed at each timestamp (see Fig. \ref{fig:waypoints} as an example). The information about the scenarios can be exported as a Matlab function to include or change the scenario variable. We can change the variables of either the ego or non-ego vehicle to simulate a potential crash scenario. We can also add a barrier or a guardrail or even another non-ego actor into the scenario to mimic the real driving situation.

\section{CONCLUSION}

In this paper, we developed a scenario-based dataset using naturalistic driving data collected around downtown Indianapolis. The dataset contains ego-vehicle waypoints, velocity, yaw angle, as well as non-ego actors' waypoints, velocity, yaw angle, entry-time, and exit-time. The real driving scenarios can be generated using the driving scenario designer in Matlab. Certain flexibility is provided to users so that actors, sensors, lanes, roads, and obstacles can be added to the existing scenarios. This dataset can be exported and it serves as a great resource for the algorithm development on motion planning and control. 

One future research direction is to enhance the existing dataset with camera-Lidar fused data. It is also interesting to study driving behavior at different traffic conditions.






\bibliography{bib}

\end{document}